\documentclass[10pt,journal,compsoc]{IEEEtran}
\usepackage{bm}
\usepackage{multirow}
\usepackage{graphicx}
\usepackage{color}
\usepackage[table,xcdraw]{xcolor}
\usepackage{amsmath,amssymb}
\usepackage[hidelinks]{hyperref}

\ifCLASSOPTIONcompsoc
  \usepackage[nocompress]{cite}
\else
  \usepackage{cite}
\fi
\ifCLASSINFOpdf
\else
\fi
\newcommand{\new}[1]{\textcolor{black}{#1}}
\newcommand{\imp}[1]{\textcolor{black}{#1}}

\begin{document}
%
\title{Video Joint Modelling Based on Hierarchical Transformer for Co-summarization}
%
%
%
%

\author{Haopeng~Li,
        Qiuhong~Ke,
        Mingming~Gong,
        and~Rui~Zhang
\IEEEcompsocitemizethanks{\IEEEcompsocthanksitem{Haopeng Li is with the School of Computing and Information Systems, University of Melbourne.  E-mail: haopeng.li@student.unimelb.edu.au.}
\IEEEcompsocthanksitem{Qiuhong Ke is with the Department of Data Science \& AI, Monash University and the School of Computing and Information Systems, University of Melbourne.  E-mail: qiuhong.ke@monash.edu.}
\IEEEcompsocthanksitem{Mingming Gong (\url{mingming-gong.github.io}) is with the School of Mathematics and Statistics, University of Melbourne. E-mail: mingming.gong@unimelb.edu.au.}
\IEEEcompsocthanksitem{Rui Zhang (\url{www.ruizhang.info}) is with Tsinghua University. E-mail: rayteam@yeah.net.}
\IEEEcompsocthanksitem{Corresponding author: Qiuhong Ke and Rui Zhang.}
}}
%
%

\markboth{Journal of \LaTeX\ Class Files,~Vol.~14, No.~8, August~2015}%
{Shell \MakeLowercase{\textit{et al.}}: Bare Advanced Demo of IEEEtran.cls for IEEE Computer Society Journals}
%



\IEEEtitleabstractindextext{%
\begin{abstract}
\imp{Video summarization aims to automatically generate a summary (storyboard or video skim) of a video, which can facilitate large-scale video retrieval and browsing.} Most of the existing methods perform video summarization on individual videos, which neglects the correlations among similar videos. Such correlations, however, are also informative for video understanding and video summarization. To address this limitation, we propose \textbf{V}ideo \textbf{J}oint \textbf{M}odelling based on \textbf{H}ierarchical \textbf{T}ransformer (\textbf{VJMHT}) for co-summarization, which takes into consideration the semantic dependencies across videos. Specifically, VJMHT consists of two layers of Transformer: the first layer extracts semantic representation from individual shots of similar videos, while the second layer performs shot-level video joint modelling to aggregate cross-video semantic information. By this means, complete cross-video high-level patterns are explicitly modelled and learned for the summarization of individual videos.
Moreover, Transformer-based video representation reconstruction is introduced to maximize the high-level similarity between the summary and the original video. Extensive experiments are conducted to verify the effectiveness of the proposed modules and the superiority of VJMHT in terms of F-measure and rank-based evaluation.
\end{abstract}


}

\maketitle

\IEEEdisplaynontitleabstractindextext

%
\IEEEpeerreviewmaketitle

\ifCLASSOPTIONcompsoc
\IEEEraisesectionheading{\section{Introduction}\label{sec:introduction}}
\else
\section{Introduction}
\fi
\IEEEPARstart{T}{he} amount of video data has been increasing exponentially on account of the popularization of online video platforms such as YouTube, Vimeo, and Facebook Watch. According to statistics in 2019, more than 500 hours of videos were uploaded to YouTube every minute\footnote{\url{https://www.tubefilter.com/2019/05/07/number-hours-video-uploaded-to-youtube-per-minute}}. To put this into perspective, it would take a person about 82 years to watch all the videos uploaded to YouTube in one hour. As a result, it is difficult to efficiently browse or retrieve useful information in the video data. To address this problem, numerous video summarization techniques have been developed in recent years \cite{mitra2016bayesian,sun2016summarizing,zhang2016video,mahasseni2017unsupervised,zhou2018deep,zhao2017hierarchical,yuan2019spatiotemporal,zhao2018hsa,fajtl2018summarizing,wei2019sequence}. Video summarization aims to automatically generate a short version of a video, which contains the important people, objects, and events in the original video. Two forms of video summary are widely exploited in previous works, \textit{i.e.}, keyframe-based 
summary and key-shot-based summary. The former picks important frames to form a static summary (\textit{i.e.}, storyboard), while the latter first segments a video into shots and then selects informative shots to form a dynamic summary (\textit{i.e.}, video skim). In this work, we aim to generate the key-shot-based summary due to its wide applications \cite{hussain2021comprehensive,li2017general,zhao2017hierarchical,zhao2018hsa,zhao2021reconstructive}.

\begin{figure}[tbp]
\centering
\includegraphics[width=0.9\columnwidth]{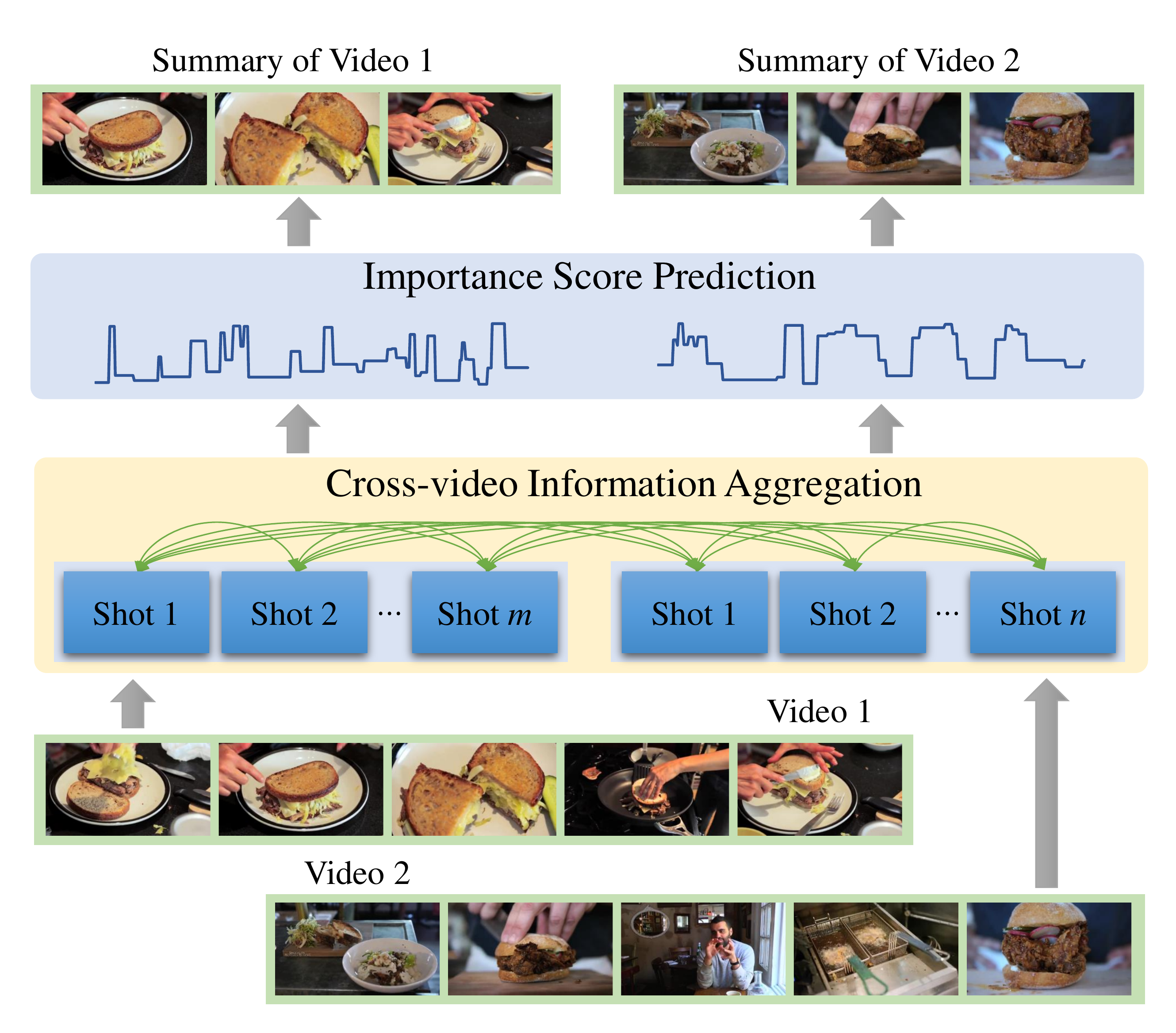}
\caption{An illustration of co-summarization for two semantically similar videos. Cross-video shot-level information aggregation is performed in our method to model the pair-wise dependencies between two arbitrary shots in the two videos. Then the shot-wise importance score is predicted based on the shot representation that contains the cross-video information. Finally, the summary of each video is generated according to the scores.}
\label{cosum}
\end{figure}

Most of the existing methods perform video summarization on individual videos, which neglects the correlations among different videos \cite{zhao2017hierarchical,zhao2018hsa,yuan2019spatiotemporal,fajtl2018summarizing}.
To resolve this issue, visual concepts across similar videos are leveraged to co-summarize the videos and improve the overall summarization performance \cite{chu2015video}. \imp{However, the method proposed in \cite{chu2015video} 1) merely finds shots that co-occur most frequently across videos to generate the summaries and 2) considers only visual similarity between two shots but neglects the important high-level information in shots.}
To address these limitations and further exploit the correlations among different videos, in this paper we propose \textbf{video joint modelling} for co-summarization, where complete cross-video high-level patterns are explicitly modelled and learned for the summarization of individual videos. Specifically, similar videos are encoded simultaneously and the semantic information is aggregated across videos in our framework.
An example of co-summarization by the proposed video joint modelling is shown in Fig. \ref{cosum}. \new{It is worth noting that the proposed video co-summarization method is not a new setting of video summarization but a new framework to address the classic video summarization task, and we perform video joint modelling only during the training of the summarization model. Specifically, during training, several semantically similar videos in the training set are encoded simultaneously, and the information is aggregated across videos. By applying supervision on all input videos, the summarization model can learn the pattern of important contents from multiple sources. After the model is trained, plain video summarization is performed on each video in the testing set individually without video joint modeling. In this case, the proposed method is comparable to existing video summarization methods.}

Intuitively, video joint modelling can be performed in the frame level, where similar videos are combined into a whole sequence and the frame-level dependencies are captured by the recurrent neural network (RNN) or Transformer \cite{vaswani2017attention}.
Therefore, video joint modelling should not be performed in the frame level plainly, and more effective network architectures are required. In this case, we develop a \textbf{hierarchical Transformer} for video joint modelling. \new{The reasons why we use Transformer instead of RNNs are: 1) RNNs struggle to capture the long-range dependencies in long sequences \cite{zhao2017hierarchical}, which greatly limits their representation ability; 2) The encoding of the current step relies on the output of the previous step in RNNs, which significantly increases the time consumption in both training and testing. In contrast, Transformer captures the global dependencies in sequences by the multi-head attention mechanism, which encodes all time steps in parallel \cite{vaswani2017attention}.} Specifically, the first layer of the hierarchical Transformer is used to extract semantic representation from individual shots of similar videos, while the second layer performs shot-level video joint modelling to aggregate cross-video semantic information. By this means, each encoded shot representation contains cross-video semantic information. \new{Since our motivation is to construct a task-specific architecture for video joint modelling, we do not modify the inner structure of the standard Transformer and we use it as the basic building block.} Besides, to retain the video-specific information, a special token is designed for each video in our framework, whose output is regarded as the video representation. Finally, the encoded shot representation (containing cross-video information) and the video representation (containing video-specific information) are combined to predict the shot-wise importance score.

In summary, we propose 
\textbf{V}ideo \textbf{J}oint \textbf{M}odelling based on \textbf{H}ierarchical \textbf{T}ransformer (\textbf{VJMHT}) for co-summarization, 
which takes into consideration the semantic dependencies across videos and the internal hierarchical structure of videos to obtain high-level video representations. Specifically, VJMHT consists of two layers of Transformer to capture the intra-shot (frame-level) and cross-video inter-shot (shot-level) dependencies hierarchically,
by which each encoded shot contains the semantic information across videos. Additionally, based on the video representation extracted by the hierarchical Transformer, we present a reconstruction loss between the representation of the video and that of its summary to maximize their similarity. 



Our contributions are summarized as follows:
\begin{itemize}
\item Video joint modelling for video co-summarization is proposed, where semantically similar videos are encoded simultaneously and the information is aggregated across videos.
\item A hierarchical Transformer is proposed for video joint modelling. By encoding the frames in each shot and the shots from all videos hierarchically, each shot representation contains high-level cross-video information.
\item Transformer-based video representation reconstruction is introduced to maximize the high-level similarity between each video and its summary. 
\end{itemize}

The rest of this paper is organized as follows. \imp{We review some related works on video summarization, inter-video communication, and Transformer in Section \ref{rw}.} Before introducing our method formally, we explain several preliminaries in Section \ref{ssr}, including the problem definition and a revisit of the standard Transformer. We then propose a novel method, \textit{i.e.}, video joint modelling based on hierarchical Transformer for co-summarization, in Section \ref{hec}. \imp{In Section \ref{exp}, we conduct extensive experiments (including comparisons with existing methods, ablation studies and sensitivity analysis) to prove the effectiveness of the proposed method.} Finally, we provide conclusions in Section \ref{con}.

\section{Related Work}
\label{rw}

\new{In this section, we first review some related work on video summarization, including the unsupervised methods and supervised ones. Then, literature on inter-video communication in various fields is summarized. Finally, since our model is constructed based on Transformer, we then give a brief introduction of the development and the applications of Transformer in computer vision (CV).}

\subsection{Video Summarization}
Various video summarization methods have been developed in previous works, which can be roughly classified into two categories: unsupervised methods and supervised methods. Each category of methods is reviewed as follows.

\subsubsection{Unsupervised Video Summarization}
The unsupervised methods focus on manually designed criteria such as the representativeness of the summary with respect to the original video \cite{de2011vsumm,ngo2003automatic} and the diversity of the frames/shots in the summary \cite{zhou2018deep}. Conventional unsupervised methods are based on classic machine learning techniques such as clustering \cite{de2011vsumm,ngo2003automatic} and dictionary learning \cite{mei20142}. VSUMM \cite{de2011vsumm} extracts color features from video frames and uses k-means to group the frames into clusters. The keyframes are determined based on the cluster centers. A unified framework is proposed in \cite{ngo2003automatic}, which employs the normalized cut algorithm to segment a video into clusters. A temporal graph is constructed based on the clusters to inherently describe the perceptual importance of video segments. Video summarization is reformulated as an $L_{2,0}$-constrained sparse dictionary selection problem in \cite{mei20142}, whose approximate solution is obtained by a simultaneous orthogonal matching pursuit (SOMP) algorithm. Recently, deep-learning-based unsupervised methods for video summarization are developed \cite{mahasseni2017unsupervised,zhou2018deep}. Adversarial learning is leveraged in \cite{mahasseni2017unsupervised}, where the discriminator is forced to classify the summary feature sequences from the video feature sequences.
DR-DSN \cite{zhou2018deep} models the video summarization task as a sequential decision-making process. The summary generator predicts the probability of a frame being selected, which is optimized by reinforcement learning in an end-to-end fashion. \new{Different from previous unsupervised methods, we use a single Transformer-based model to extract semantic representations from videos and summaries. The distance between the representations of generated summaries and their corresponding videos is used as the loss to train the model. Our method has the following advantages with respect to previous ones: a) The proposed framework consists of only one model (hierarchical Transformer) to generate summaries and perform video representation reconstruction; b) The optimization of our model is much more efficient since no adversarial learning or reinforcement learning is involved.}

\subsubsection{Supervised Video Summarization}
RNN-based models dominate the supervised methods for video summarization in recent years. A number of variants of RNN are proposed to encode and summarize the video from various aspects. dppLSTM \cite{zhang2016video} uses bi-directional LSTM to model the temporal dependencies in the video. 
Considering the hierarchical structure of videos, hierarchical LSTM is developed in \cite{zhao2017hierarchical}, which uses two layers of LSTM to model the intra-shot dependencies and inter-shot dependencies. Furthermore, structure-adaptive hierarchical LSTM is adopted in \cite{zhao2018hsa}, where shot boundary detection is performed as an extra task. 
Instead of using LSTM to model videos, SUM-FCN \cite{rochan2018video} captures the long-range dependencies within a video by a one-dimension fully convolutional network (FCN). By a stack of convolutions and pooling layers, the effective context size grows as it goes deeper in the network. Considering the deficiencies of the recurrent models, VASNet \cite{fajtl2018summarizing} introduces the attention mechanism to capture the global dependencies among video frames. Video summarization is reformulated as a sequence-to-sequence task in \cite{ji2019video}, where an LSTM-based encoder-decoder network with an intermediate attention layer is leveraged. Besides considering the temporal dependencies, a three-dimension convolutional neural network (3D-CNN) is used to extract spatial-temporal features of videos in \cite{yuan2019spatiotemporal}, and the temporal dependencies are further modelled by the recurrent network. A novel loss function called Sobolev loss is proposed in \cite{yuan2019spatiotemporal} to capture the local dependencies. Spatiotemporal features are extracted to generate the inter-frames motion curve in \cite{huang2019novel}, which is used for video segmentation. Then a self-attention model is employed to select keyframes inside the shots. \new{SumGraph \cite{park2020sumgraph} is a graph-based video summarization method, where the global dependencies of frames are modeled by recursive graph modeling networks. Despite that they achieve great success in video summarization, previous methods have the limitation of considering only the correlations among frame/shots within individual videos. To address this issue, we propose video joint modelling to capture the correlations among semantically similar videos for video summarization. To perform video joint modelling, a novel hierarchical Transformer is proposed to aggregate shot-level information across videos. Different from previous supervised methods, the proposed model not only considers the inner structure videos (frame-shot-video), but also captures the global dependencies within individual videos and across videos.}

\subsection{\new{Inter-video Communication}}

\new{Inter-video communication has been exploited in various computer vision fields such as action localization \cite{zhang2021i2net,wang2021exploring}, video object detection \cite{han2020mining,wang2020contrastive}, and video person re-identification \cite{zhang2019scan}. Inter-Video Proposal Relation module is proposed in \cite{han2020mining} to learn robust object representations by capturing the correlations of hard samples. Self-and-Collaborative Attention Network (SCAN) \cite{zhang2019scan} takes a pair of videos as input and measures their similarity to perform person re-identification. In SCAN, the attention mechanism is leveraged to refine the representations of videos by considering both the intra-sequence and inter-sequence correlations. In \cite{wang2020contrastive}, correspondence estimation is achieved by simultaneously modelling both the intra-video and inter-video representation associations for self-supervised learning. The learned representations are successfully applied to video object segmentation and tracking. I2Net \cite{zhang2021i2net} exploits both the intra-video and inter-video attention to address the issues of lacking long-term relationships and action pattern uncertainty in temporal action localization. Cross-video relationships are mined for weakly supervised temporal action localization in \cite{wang2021exploring}, where cross-video modelling is performed in the sub-action granularity. Different from previous works, we proposed shot-level cross-video information aggregation which not only considers the internal structure of videos, but also decreases the computational complexity when dealing with long videos.}

\subsection{Transformer in CV}

Transformer has revealed great power in computer vision tasks such as object detection and action recognition. Object detection is modelled as a sequential-to-sequential task and addressed using Transformer in \cite{carion2020end} and \cite{zhu2020deformable}. They do not need many hand-designed components like anchor generation and non-maximum suppression. Transformer is also exploited for the recognition of actions of individuals or groups in \cite{girdhar2019video} and \cite{gavrilyuk2020actor} by spatial-temporal context aggregation. \new{In this paper, we propose hierarchical Transformer to model the dependencies of frames within each shot and those of shots among different videos successively for video co-summarization.}


\section{Preliminaries}
\label{ssr}

In this section, we first introduce the task of video summarization, including the purpose and the formulation. We then revisit the standard Transformer since our model is constructed based on it.

\subsection{Video Summarization}

Video summarization aims to automatically generate a short version of a video, which contains the important people, objects, and events. By skimming through only the summary, a viewer can infer the main content of the original video. \new{In most previous works \cite{zhang2016video,rochan2018video,zhang2018retrospective} and our paper, video summarization is defined as a frame-wise binary classification task. Fully consistent with previous works, the aim and process of video summarization are described in an abstract and mathematical fashion as follows.} Formally, given a video sequence $\left\lbrace \bm{X}_j\right\rbrace_{j=1}^M$ ($M$ is the number of frames and $\bm{X}_j\in\mathbb{R}^{H\times W \times 3}$ is the $j$-th frame), video summarization ($\mathrm{VS}$) maps the sequence to a binary sequence $\{y_j\}_{j=1}^M$ ($y_j\in\{0,1\}$) whose elements indicate whether frames are selected into the summary, \textit{i.e.}, 
\begin{equation}
\mathrm{VS}\left(\left\lbrace \bm{X}_j\right\rbrace_{j=1}^M\right)=\{y_j\}_{j=1}^M,
\end{equation}
where $y_j=1$ means the $j$-th frame is selected into the summary, while $j=0$ means the opposite. In this case, the video summary is formed as
\begin{equation}
S=\{\bm{X}_j|y_j=1,j=1,2,\cdots,M\}.
\end{equation}
\new{Note that, in most video summarization methods \cite{zhang2016video,zhao2018hsa,ji2019video,zhao2021reconstructive}, there is a limit to the length of the video summary, \textit{i.e.}, $|S|\leq \gamma M$, where $|S|$ is the number of frames in $S$, and $\gamma$ is a pre-defined ratio. Such constraint requires the video summarization methods generate summaries which contain more important frames and less redundancy.}

\subsection{Standard Transformer Revisit}

In this section, we revisit the standard Transformer, which is the crucial component in our model. Transformer is a sequence-to-sequence model consisting of an encoder and a decoder, each of which is elaborated as follows.

\begin{figure}[tbp]
\centering
\includegraphics[width=\columnwidth]{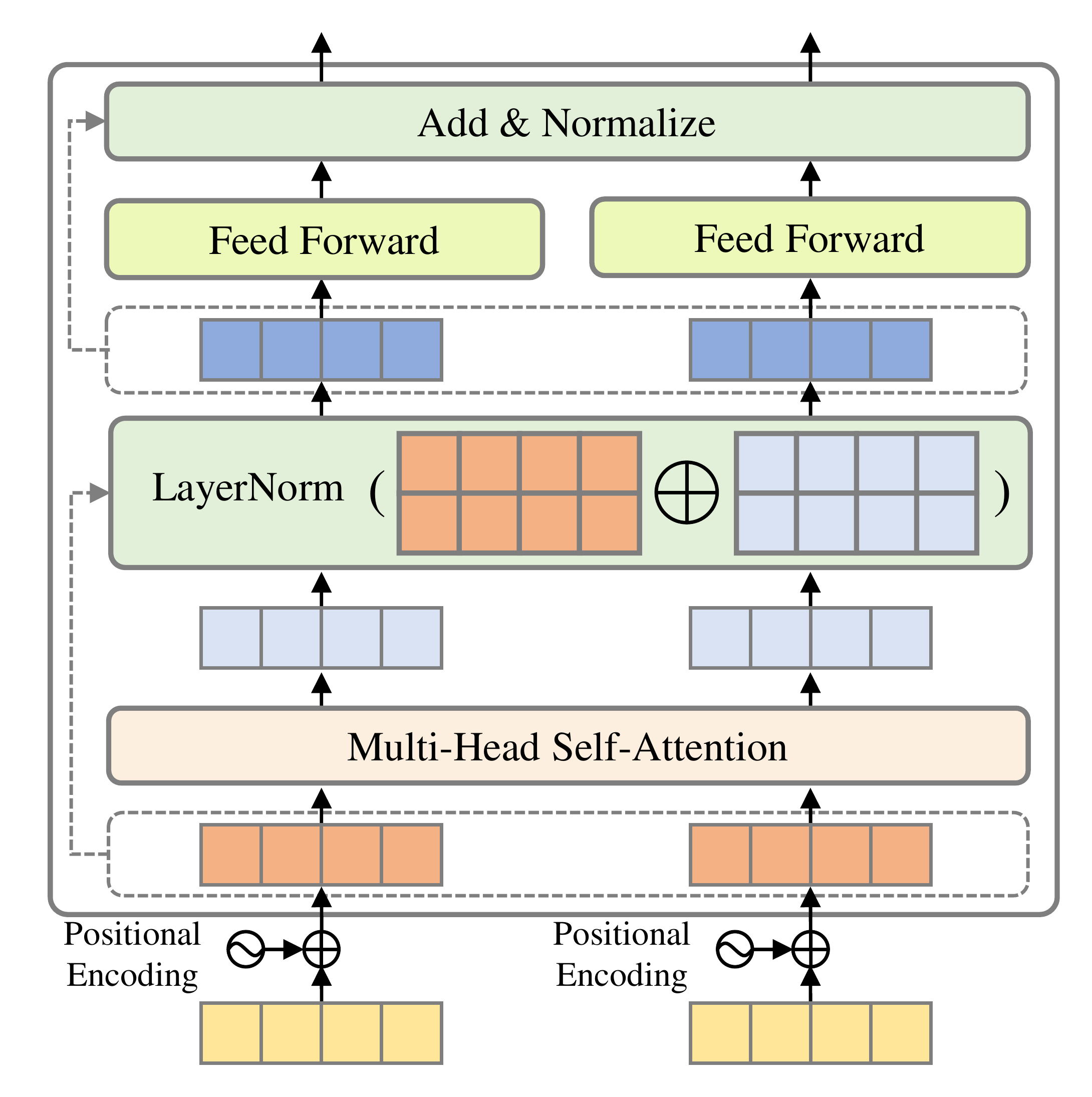}
\caption{The structure of the first layer of Transformer encoder. For simplicity, only two tokens are shown. The positional encodings are added to the input sequence at the beginning. In each layer, the self-attention mechanism is applied to the sequence, followed by a residual connection and layer normalization. Then, a positional-wise feed-forward network is used to transform each feature into a hidden space. At last, a residual connection and layer normalization are equipped.}
\label{trans}
\end{figure}

The Transformer encoder takes as input a sequence of feature vectors and outputs a sequence of encoded features. The output is usually referred to as the memory of the Transformer encoder \cite{vaswani2017attention,carion2020end}. Specifically, the encoder is composed of a stack of identical layers. Each layer has two sub-layers: a multi-head self-attention mechanism and a position-wise fully connected feed-forward network. Each sub-layer is equipped with a residual connection \cite{he2016deep} and layer normalization
\cite{ba2016layer}. The structure of the first layer of Transformer encoder is shown in Fig. \ref{trans}. Formally, given an input sequence $\bm{X}\in \mathbb{R}^{N\times d}$, where $N$ is the number of time-steps of the sequence and $d$ is the feature dimension, the outputs of the two sub-layers (denoted as $\bm{X}'\in \mathbb{R}^{N\times d}$ and $\bm{X}''\in \mathbb{R}^{N\times d}$) 
can be expressed as:
\begin{align}
&\bm{X}' = \mathrm{LayerNorm}(\mathrm{MultiHead}(\bm{X},\bm{X},\bm{X})+\bm{X}), \\
&\bm{X}'' = \mathrm{LayerNorm}(\mathrm{FFN}(\bm{X}')+\bm{X}'),
\end{align}
where $\mathrm{MultiHead}(\cdot,\cdot,\cdot)$ is the multi-head attention mechanism, $\mathrm{LayerNorm}(\cdot)$ is the layer normalization, and $\mathrm{FFN}(\cdot)$ is the position-wise fully connected feed-forward network.

\begin{figure*}[tbp]
\centering
\includegraphics[width=\textwidth]{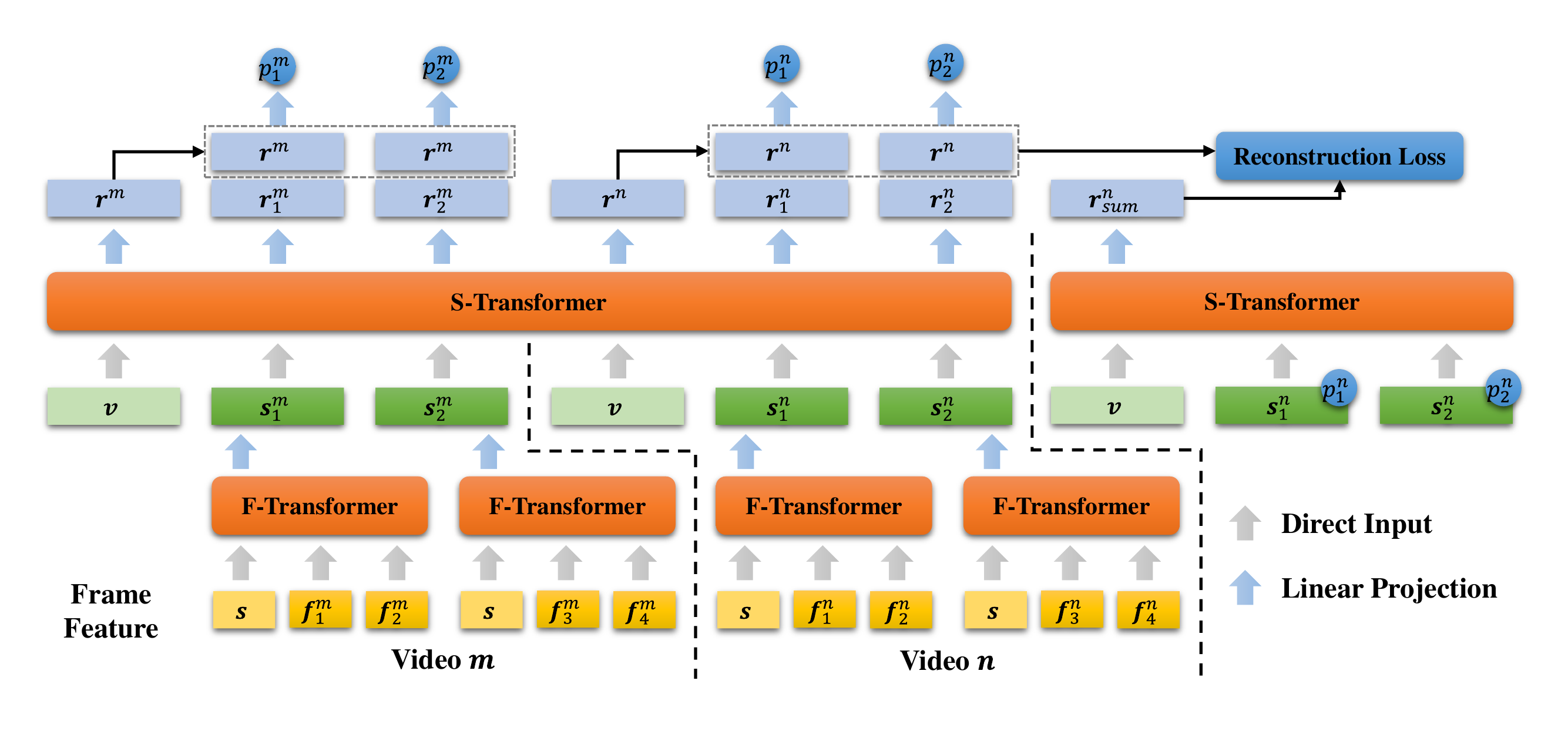}
\caption{The overview of video joint modelling based on hierarchical Transformer (VJMHT) for co-summarization. \imp{Without loss of generality, only two videos are shown, and we assume each video consists of two shots and each shot consists of two frames.} Two semantically similar videos are sent into VJMHT simultaneously. The frames in each shot are encoded by F-Transformer in parallel to obtain the shot embedding. The shots are aggregated across videos in S-Transformer. The encoded shot embeddings along with the video representation are combined to predict the shot-wise importance scores. In addition, video representation reconstruction is performed to minimize the distance between the representation of the summary and the video. Transformers/linear projections in the same level have the same structure and share parameters.}
\label{overview}
\end{figure*}

The multi-head attention mechanism is a 
crucial component in Transformer, which takes as input a query matrix $\bm{Q}$, a key matrix $\bm{K}$ and a value matrix $\bm{V}$ ($\bm{X}=\bm{Q}=\bm{K}=\bm{V}$ in the multi-head self-attention), and captures global dependencies using scaled dot-product attention in different subspaces. Formally,
\begin{equation}
\mathrm{MultiHead}(\bm{Q},\bm{K},\bm{V})=\mathrm{Concat}\left(\bm{H}_1,\cdots,\bm{H}_h\right)\bm{W}^o,
\end{equation}
where
\begin{equation}
\bm{H}_i=\mathrm{softmax}\left(\frac{\bm{Q}\bm{W}^q_i\left(\bm{K}\bm{W}^k_i \right)^\mathrm{T}}{\sqrt{d_h}}\right)\bm{V}\bm{W}^v_i,
\end{equation}
\imp{where $h$ is the number of heads. 
$d_h=d/h$. 
$\bm{W}^q_i$, $\bm{W}^k_i$,$\bm{W}^v_i\in\mathbb{R}^{d\times d_h}$ and
$\bm{W}^o\in\mathbb{R}^{d\times d}$ are parameters. 
By transforming the concatenated attention outputs from all subspaces, the model attends to information from different aspects.}

The position-wise fully connected feed-forward network consists of two fully connected layers with a ReLU activation in between, which transforms the feature in each position separately and identically, \textit{i.e.},
\begin{equation}
\mathrm{FFN}(\bm{X}')=\max\left\lbrace 0,\bm{X}'\bm{W}_1+\bm{b}_1\right\rbrace\bm{W}_2+\bm{b}_2, 
\end{equation}
\imp{where $\bm{W}_1\in\mathbb{R}^{d\times d'},\bm{W}_2\in\mathbb{R}^{d'\times d},\bm{b}_1\in\mathbb{R}^{d'},\bm{b}_2\in\mathbb{R}^{d}$ are learnable parameters}. Note that the positional encodings (learned or fixed) are first added to the input features of the first layer to make use of the order information. 

In terms of the Transformer decoder, it is similar to the encoder except that in each layer, an extra multi-head cross-attention mechanism is positioned after the multi-head self-attention mechanism. In our model, only the Transformer encoder is used to encode the videos in a hierarchical fashion, while the decoder is discarded. For the prediction of importance of video shots, a simple linear projection is applied after the encoder. In this case, Transformer refers to the Transformer encoder throughout this paper.

\section{The Proposed Method}
\label{hec}

In this section, we elaborate Video Joint Modelling based on Hierarchical Transformer (VJMHT) for co-summarization. The overview of VJMHT is illustrated in Fig. \ref{overview}. Specifically, the structure of the proposed framework is first explained, including intra-shot dependency modelling and cross-video inter-shot dependency modelling. We then perform Transformer-based video representation reconstruction to maximize the similarity between the summary and the video. Finally, we explain the optimization of VJMHT and the details of inference.

\subsection{Video Joint Modelling for Co-summarization}
\label{si}

Conventional video summarization methods generate summaries of videos separately,
without considering the connections between videos. However, different videos of the similar topic may contain the same important information which can be leveraged to improve the the summary of each video \cite{chu2015video}.
In this paper, based on the hierarchical structure, video joint modelling is developed for video co-summarization. The proposed framework consists of two layers of Transformer: F-Transformer for intra-shot (frame-level) dependencies capturing and S-Transformer for cross-video inter-shot (shot-level) dependencies capturing, each of which is explained as follows.


\subsubsection{Intra-shot Dependency Modelling}
Since the intra-shot dependency modelling for each video is independent and identical, we explain this process using only one video.
\imp{Given a video sequence
$V=\left\lbrace \bm{f}_j\right\rbrace_{j=1}^M$ ($M$ is the number of frames and $\bm{f}_j\in\mathbb{R}^{d_f}$ is the feature of the $j$-th frame), the shot boundary detection algorithm, Kernel-based Temporal Segmentation (KTS) \cite{potapov2014category}, is first applied to obtain the shot boundaries $\left\lbrace b_i\right\rbrace_{i=0}^P$, where $P$ is the number of shots, and $b_0=1,b_P=M$ represent the indexes of the first frame and the last one.} In this case, the $i$-th shot $V_i$ is a sub-sequence of $V$, \textit{i.e.}, $V_i=\left\lbrace \bm{f}_{b_{i-1}+1},\bm{f}_{b_{i-1}+2},\cdots, \bm{f}_{b_{i}}\right\rbrace\subseteq V$. For the sake of clarity, \textbf{$\bm{j}$ and $\bm{i}$ are used as the index of frame and that of shot, respectively, throughout this paper.} Since the intra-shot dependency modelling aggregates information within each shot $V_i$, we explain this process using only one shot as follows.

\imp{For each shot in a video, inspired by the \texttt{[CLS]} token in BERT \cite{devlin2018bert}, a learnable shot embedding $\bm{s}\in\mathbb{R}^{d_f}$ is randomly initialized and prepended to the frames in the shot, whose state at the output of F-Transformer serves as the shot representation.} Fixed sinusoidal positional encodings \cite{vaswani2017attention} are added to the frame features to retain 
the order information. Formally, given a shot starting at the $t$-th frame and consisting of $L$ frames  $\left\lbrace \bm{f}_j\right\rbrace_{j=t}^{t+L-1}$, the input of F-Transformer is
\begin{equation*}
\bm{F}=\left[\bm{s};\bm{f}_t;\bm{f}_{t+1};\cdots;\bm{f}_{t+L-1} \right]+\bm{E}^f_{pos}\in\mathbb{R}^{(1+L)\times d_f},
\end{equation*}
where $\bm{E}^f_{pos}$ is the positional encoding matrix for frames. After being encoded by F-Transformer, the dependencies within the shot are retained in the memory of F-Transformer.
As stated above, the first embedding in the memory is regarded as the representation of the shot. Assuming that a video consists of $P$ shots, by encoding $P$ shots in parallel, a sequence of shot representations $\left\lbrace \bm{s}_i\right\rbrace_{i=1}^{P}$ ($\bm{s}_i\in\mathbb{R}^{d_s}$) is obtained, where the dimension transformation is performed by linear projection ($\mathbb{R}^{d_f}\rightarrow\mathbb{R}^{d_s}$). $\left\lbrace \bm{s}_i\right\rbrace_{i=1}^{P}$ is sent into S-Transformer for shot-level modeling.

Note that in F-Transformer, all the videos are encoded in parallel, which means the information is not aggregated across videos. Frame-level aggregation is not performed for two reasons: 1) The quadratic time and memory complexity with respect to sequence length seriously restricts Transformer from processing all frames in all videos as a whole sequence \cite{tay2020efficient}; 2) There exists a great deal of redundancy in frame-level modelling since adjacent frames are visually and semantically similar;
3) Our purpose is to capture the shot-level semantic similarity among videos, while individual frames are not capable of representing such complex concepts. In this case, video information aggregation is only performed in S-Transformer as explained in the next part.

\subsubsection{Cross-video Inter-shot Dependency Modelling}

Different to the intra-shot dependency modelling that is independent and identical for each video, the inter-shot modelling involves cross-video information aggregation. Given $N$ semantically similar videos in the training set (the similarity between videos is explained in Section \ref{set}), they are encoded by F-Transformer independently as explained above, and the shot representations of the $N$ videos are denoted as $\left\lbrace \bm{s}^1_i\right\rbrace_{i=1}^{P^1},\left\lbrace \bm{s}^2_i\right\rbrace_{i=1}^{P^2},\cdots,\left\lbrace \bm{s}^N_i\right\rbrace_{i=1}^{P^N}$, where $\left\lbrace P^n\right\rbrace_{n=1}^N$ are the shot numbers of $N$ videos. \textbf{Superscripts are used as the video indexes throughout this paper.} 

Since $\left\lbrace \bm{s}^1_i\right\rbrace_{i=1}^{P^1},\left\lbrace \bm{s}^2_i\right\rbrace_{i=1}^{P^2},\cdots,\left\lbrace \bm{s}^N_i\right\rbrace_{i=1}^{P^N}$ represent the semantic information of shots in $N$ videos, they are sent into S-Transformer simultaneously to perform information aggregation across videos. Formally, the input of S-Transformer is expressed as 
\begin{equation*}
\bm{S}=\left[\bm{v};\bm{s}^1_1;\cdots;\bm{s}^1_{P^1};\bm{v};\bm{s}^2_1;\cdots;\bm{s}^2_{P^2};\bm{v};\bm{s}^N_1;\cdots;\bm{s}^N_{P^N} \right]+\bm{E}^s_{pos}.
\end{equation*}
Note that an identical learnable video embedding $\bm{v}\in\mathbb{R}^{d_s}$ is prepended to the shot embeddings of each video. In addition, \textbf{a mask is applied to restrict each $\bm{v}$ from attending to shots of other videos}. Hence, the video embeddings can only aggregate information from shots of their corresponding videos, while the dependencies among all shot embeddings are captured.
An extra linear projection ($\mathbb{R}^{d_s}\rightarrow\mathbb{R}^{d_v}$) is applied to the output of S-Transformer to obtain the the co-representation of the $N$ videos. The transformed representation is denoted as 
\begin{equation*}
\bm{R}=\left[\bm{r}^1;\bm{r}^1_1;\cdots;\bm{r}^1_{P^1};\bm{r}^2;\bm{r}^2_1;\cdots;\bm{r}^2_{P^2};\bm{r}^N;\bm{r}^N_1;\cdots;\bm{r}^N_{P^N} \right],
\end{equation*}
where $\left\lbrace \bm{r}^n\right\rbrace_{n=1}^{N}$ are the representations of $N$ videos, containing the video-specific information, while $\left\lbrace \bm{r}^n_i\right\rbrace_{i=1}^{P^n}$ are the representations of the shots in the $n$-th video, containing the cross-video information. $\left\lbrace \bm{r}^n\right\rbrace_{n=1}^{N}$ and $\left\lbrace \bm{r}^n_i\right\rbrace_{i=1}^{P^n}$ are used for video co-summarization as shown in Fig. \ref{overview}.

To predict the importance scores, each encoded shot representation $\bm{r}^n_i$ is concatenated with the encoded representation of its corresponding video $\bm{r}^n$. In this way, the concatenated representation contains not only the information across shots from $N$ videos, but also the specific content of the $n$-th video. Finally, the concatenated representation is used to predict the score of the $i$-th shot in the $n$-th video by linear projection, \textit{i.e.},
\imp{\begin{equation}
p_i^n=\mathrm{Concat}(\bm{r}^n_i,\bm{r}^n)\bm{W}_p+b_p,
\end{equation}
where $\bm{W}_p\in\mathbb{R}^{2d_v},b_p\in\mathbb{R}$ are the learnable projection parameters.} For frame-wise importance score, each frame is assigned with the score of the shot it belongs to. The score the $j$-th frame in the $n$-th video is denoted as $\bar{p}_j^n$.

\subsection{Transformer-based Video Representation Reconstruction}
\label{rec}

A satisfactory video summary is supposed to contain the main content of the original video. This property is referred to as reconstruction capacity, which is exploited to enhance the performance of video summarization \cite{mahasseni2017unsupervised,zhou2018deep}. However, most of the existing methods perform reconstruction using the RNN-based representations, which is limited since: 1) RNNs struggle to capture the long-term dependencies in videos; 2) the structural information of videos is not retained. To address these issues, we propose Transformer-based video representation reconstruction in this work.
Since the video representations extracted by the proposed hierarchical Transformer encode the main content and the structure of the videos, it is utilized for the high-level video reconstruction in our method.

More specifically, by co-summarization, the score of the $i$-th shot in the $n$-th video $p_i^n$ is obtained,
which is regarded as the weight of the shot to form the summary. In this case, the shot embeddings of the summary are computed as the weighted shot embeddings of the original video, \textit{i.e.},
\begin{equation*}
\left\lbrace p_i^1\bm{s}^1_i\right\rbrace_{i=1}^{P^1},\left\lbrace p_i^2\bm{s}^2_i\right\rbrace_{i=1}^{P^2},\cdots,\left\lbrace p_i^N\bm{s}^N_i\right\rbrace_{i=1}^{P^N},
\end{equation*}
each of which is sent into S-Transformer independently. Note that, the video joint modelling is not performed when encoding the shots of summaries. Following the process in Section \ref{si}, the representations of $N$ summaries are obtained by the S-Transformer and denoted as $\left\lbrace \bm{r}^n_{sum}\right\rbrace_{n=1}^N$ as shown in Fig. \ref{overview}. \imp{Since $\bm{r}^n_{sum}$ encodes the content of the $n$-th summary, it is supposed to be close to $\bm{r}^n$ which encodes the content of the $n$-th video. In this case, a Transformer-based video representation reconstruction loss is proposed,}
\begin{equation}
\mathcal{L}_{rec}=\frac{1}{d_vN}\sum_{n=1}^N{\|\bm{r}^n_{sum}-\bm{r}^n\|_2^2},
\end{equation}
where $d_v$ is the dimension of the video representations.

\begin{table*}[tbp]
\caption{A summary of the datasets used in the experiments.}
\label{ds}
\center
\begin{tabular}{cccccc}
\hline
Dataset & \# of Videos & Duration (min) & Genres/Topics                           & Annotations                & Usage              \\ \hline
SumMe \cite{gygli2014creating}   & 25           & 1--6           & Holidays, cooking and sports        & 15--18  sets of key-shots  & Training\&Testing \\
TVSum \cite{song2015tvsum}   & 50           & 2--10          & Beekeeping, parade and dog show & 20 set  shot-level  scores & Training\&Testing \\ \hline
YouTube \cite{de2011vsumm} & 50           & 1--10          & Sports, news and TV-shows     & 5  sets of key-frames      & Training         \\
OVP \cite{de2011vsumm}     & 39           & 1--4           & Documentaries & 5  sets of key-frames      & Training         \\
 \hline
\end{tabular}
\end{table*}
\subsection{Optimization and Inference}

The proposed architecture is optimized in an end-to-end fashion. Specifically, the loss function $\mathcal{L}$ is the linear combination of three terms: the supervision loss $\mathcal{L}_{sup}$, the reconstruction loss $\mathcal{L}_{rec}$ (as proposed in Section \ref{rec}) and the regularization $\mathcal{L}_{reg}$, \textit{i.e.},
\begin{equation}
\label{ul}
\mathcal{L}=\mathcal{L}_{sup}+\alpha\mathcal{L}_{rec}+\beta\mathcal{L}_{reg},
\end{equation}
where $\alpha,\beta$ are hyper-parameters to balance the three terms. For each video, the supervision loss is the mean squared error (MSE) between the predicted frame-wise importance scores and the ground truth ones, \textit{i.e.},
\begin{equation}
\mathcal{L}_{sup}=\frac{1}{M}\sum_{j=1}^M{\left(\bar{p}_j-g_j\right)^2},
\end{equation}
where $M$ is the number of frames, and $\left\lbrace \bar{p}_j\right\rbrace_{j=1}^M,\left\lbrace g_j\right\rbrace_{j=1}^M$ are the predicted frame importance scores and the ground truth ones, respectively. Following \cite{mahasseni2017unsupervised,zhou2018deep}, we introduce the regularization to prevent the model from predicting high scores for all frames indiscriminately. Specifically, for each video, the regularization is computed as follows,
\begin{equation}
\label{reg}
\mathcal{L}_{reg}=\left(\frac{1}{M}\sum_{j=1}^M{\bar{p}_j}-\varepsilon \right)^2,
\end{equation}
where $\varepsilon$ is a hyper-parameter controlling the percentage of frames to be selected into the summary. The proposed model can also be trained in an unsupervised manner, where $\mathcal{L}_{sup}$ in Eq. (\ref{ul}) is discarded. \imp{We also evaluate the unsupervised VJMHT in the experiments.}

To generate a summary, shots are selected to maximize the total score while ensuring the length of the summary is less than a pre-defined limit, which is set to 15\% of the original video as in \cite{zhang2016video,mahasseni2017unsupervised,zhou2018deep}. The maximization is modelled as the 0/1 Knapsack problem \cite{song2015tvsum}, which is solved by dynamic programming.

\new{Note that the proposed video joint modelling is performed only during training, and the testing process is the same as plain video summarization. Specifically, during training, several semantically similar videos in the training set are encoded simultaneously, and the information is aggregated across videos. By applying supervision on all input videos, the summarization model can learn the pattern of important contents from multiple sources. After the model is trained, plain video summarization is performed on each video in the testing set individually without video joint modeling. The reasons why we do not perform video co-summarization during testing are: 1) Finding semantically similar videos for the testing videos is not practical, as no prior information about the testing videos is available in practical applications; 2) Video co-summarization requires much more computational resource compared with plain video summarization, so the efficiency of testing cannot be guaranteed if video joint modelling is applied; 3) As the topic prior of videos is used in co-summarization, we perform plain video summarization instead of co-summarization to ensure fair comparisons with other methods. 4) The trained model is capable of modelling the complete high-level patterns of important contents, so it is also effective when being applied to single-video summarization. Additionally, compared with traditional training, the training with video joint modelling is supervised by multiple correlated samples. Such training manner renders the model robust to complicated videos and capable of extracting powerful features for video summarization during testing.}

\section{Experiments}
\label{exp}

In this section, we conduct extensive experiments to veriry the effectiveness of the proposed methods. First, we elaborate the experiment settings, including the datasets, the implementation details and the evaluation metrics. \imp{To demonstrate the superiority of the proposed VJMHT, we compare it with existing methods in terms of F-measure and rank correlation coefficients.} We then conduct ablation studies to illustrate the impact of crucial components in our model. Besides, we visualize the summary generated by our methods as well as the importance score output by VJMHT. Finally, we conduct sensitivity analysis on the structure parameters of the proposed model.

\subsection{Experiment Settings}
\label{set}

\subsubsection{Datasets}
Two public benchmarks for video summarization are used to evaluate the proposed method: SumMe \cite{gygli2014creating} and TVSum \cite{song2015tvsum}. SumMe consists of 25 videos with a coverage of various events such as holidays and cooking. The length of the videos in SumMe varies from 1.5 min to 6.5 min. Each video in SumMe was annotated by 15--18 persons with key shots as summaries. TVSum has 50 videos of 10 categories (\textit{e.g.}, Beekeeping and Parade) from TRECVid Multimedia Event Detection (MED) \cite{smeaton2006evaluation}. The length of the videos in TVSum varies from 2 min to 10 min. Similarly, each video was annotated by 20 persons with frame-wise importance scores. Additionally, YouTube \cite{de2011vsumm} and OVP\footnote{Open video project: \url{https://open-video.org}} are used for training. The videos in YouTube cover a variety of topics such as sports and news, while the videos in OVP are mostly documentaries. YouTube has 39 videos (cartoons are excluded) and OVP has 50 videos, which were annotated with keyframe-based summaries. The keyframes are converted to importance scores for training as in \cite{zhang2016video}. A summary of the datasets used in the experiments is shown in TABLE \ref{ds}.

\new{Following previous works on video summarization \cite{zhao2021reconstructive,jung2019discriminative,rochan2018video,zhang2018retrospective}, we compare our methods with the state of the arts in three settings: the canonical setting, the augmented setting, and the transfer setting. Specifically, the canonical setting is the standard supervised learning setting where the training set and the testing set are subsets of one dataset. Considering that SumMe and TVSum are relatively small for training, the augmented setting is adopted to augment the training samples. In this setting, the videos in the YouTube dataset and the OVP dataset are added into the training set, and the testing set remains the same. As for the transfer setting, it is exploited to measure the transfer ability of the summarization model, where the model is trained on SumMe (TVSum) and tested on TVSum (SumMe). Note that the training samples are also augmented with the videos in YouTube and OVP.}

\subsubsection{Implementation Details}
\label{id}
\noindent \textbf{Pre-processing and Feature Extraction.} Following \cite{zhang2016video}, each video is sub-sampled to 2 frames per second to remove redundancy. The output of the penultimate layer (pool 5) of the  GoogLeNet \cite{szegedy2015going} pre-trained on ImageNet \cite{russakovsky2015imagenet} is adopted as the frame feature ($d_f=1024$). 

\noindent \textbf{Network Structure.}
Since F-Transformer and S-Transformer encode frames and shots respectively, they have different structures. Specifically, in F-Transformer, 2 identical layers are stacked and the dimension of the feed-forward network is set to 4,096 in all layers. The dimensions of the shot embedding and video embedding are set to 512 ($d_s=d_v=512$). In S-Transformer, 3 identical layers are stacked and the dimension of the feed-forward network is set to 2,048 in all layers. The number of heads in the multi-head attention is set to 2 in both F-Transformer and S-Transformer. Dropout is discarded in VJMHT. \imp{Hyper-parameter sensitivity analysis is conducted in Section \ref{fa}.}

\noindent \textbf{Video Joint Modelling.}
Two semantically similar videos are sent into VJMHT simultaneously considering the memory limitation ($N=2$), and the video reconstruction is performed on only one of them. 
\new{The details of the clustering process, training, and testing are explained as follows. Firstly, we pretrain the hierarchical Transformer on YouTube and OVP. Then, we use the pretrained Transformer to extract semantic representations of the training videos in SumMe, TVSum, YouTube, and OVP. After obtaining the video representations, we utilize K-Means to group all the training videos into 25 clusters. By this means, the videos in the training set of SumMe/TVSum are divided into several groups based on the above clusters. Taking the training on TVSum in the canonical setting for example, several videos are sampled from the same group and sent into the proposed framework for summarization. After the model is trained, the videos in the testing set of TVSum are sent into the model individually for regular summarization. The same goes for SumMe and other settings. In this case, no additional prior information of videos is used in the testing phase. Therefore, all the comparisons with other methods in our experiments are fair.}

\noindent \textbf{Training.}
\imp{$\alpha,\beta$ in Eq. (\ref{ul}) are set to 0.01 and 0.1, respectively.} $\varepsilon$ in the regularization is set to 0.5. VJMHT is optimized by Adam \cite{kingma2014adam} for 60 epochs with batch size 1 and initial learning rate $10^{-5}$, which is reduced to $10^{-6}$ at the 30th epoch. Since SumMe and TVSum are relatively small, five-fold cross-validation is performed to make a fair comparison as in existing methods \cite{he2019unsupervised,jung2019discriminative,zhou2018deep}. All experiments are conducted using PyTorch on NVIDIA Tesla P100 GPUs.

\subsubsection{Evaluation Metrics}
In this paper, the summaries are evaluated by two kinds of metrics: F-measure and rank correlation coefficients (Kendall’s $\tau$ and Spearman’s $\rho$).

\begin{table*}[htbp]
\caption{\imp{The results (F-measure) in different settings.} The methods above VJMHT$_{\bm{uns}}$ are unsupervised methods or weakly supervised ones, while the methods below VJMHT$_{\bm{uns}}$ are supervised methods. The best and the second-best results are in \textbf{bold} and \underline{underlined}, respectively.}
\label{res1}
\center
\begin{tabular}{l|ccc|ccc}
	\hline
	\multirow{2}{*}{Methods} & \multicolumn{3}{c|}{SumMe} & \multicolumn{3}{c}{TVSum} \\
	& Canonical & Augmented & Transfer & Canonical & Augmented & Transfer\\ \hline
	SUM-GAN \cite{mahasseni2017unsupervised}& 0.387  & 0.417 & ---  &  0.508 & 0.589 & --- \\
	DR-DSN \cite{zhou2018deep}& 0.414  & 0.428  & 0.424  & 0.576  & 0.584  & 0.578  \\
	SUM-FCN$_{uns}$ \cite{rochan2018video}& 0.415& --- &--- &0.527 &---  & --- \\
	PCDL \cite{zhao2019property}& 0.427  & ---  & ---  & 0.584  & ---  &---\\
	ACGAN \cite{he2019unsupervised}&0.460 & 0.470  &\underline{0.445} &0.585 & 0.589 & 0.578 \\
	UnpairedVSN \cite{rochan2019video}  & 0.475&  --- & ---&0.556 &  ---& --- \\
	WS-HRL \cite{chen2019weakly}&0.436 & 0.445  &--- & 0.584 &0.585  & --- \\
	RSGN$_{uns}$ \cite{zhao2021reconstructive}&0.423&0.436&0.412&0.580&0.591&\underline{0.597}\\\hline
	VJMHT$_{uns}$ (Ours) &0.471 &  0.490& 0.436 & 0.573 & 0.597 & 0.571 \\ \hline
	dppLSTM \cite{zhang2016video}& 0.386  & 0.429  & 0.418  & 0.547  &  0.596 &  0.587\\
	SUM-GAN$_{sup}$ \cite{mahasseni2017unsupervised}& 0.417 & 0.436  &---  &0.563  &0.612  & ---   \\
	H-RNN \cite{zhao2017hierarchical}& 0.421 & 0.438  & --- & 0.579 & \underline{0.619} &  ---  \\
	HSA-RNN \cite{zhao2018hsa} & 0.423 & 0.421  & --- & 0.587 & 0.598 &  ---  \\
	re-SEQ2SEQ \cite{zhang2018retrospective} & 0.425 & 0.449  & --- & \underline{0.603} & \textbf{0.639} & ---   \\
	SUM-FCN \cite{rochan2018video}& 0.475 & \underline{0.511} & 0.441 & 0.568 &0.592 & 0.582  \\
	CSNet$_{sup}$ \cite{jung2019discriminative}&0.486 &  0.487 & 0.441&0.585 & 0.571 & 0.574 \\ 
	ACGAN$_{sup}$ \cite{he2019unsupervised}& 0.472& ---  &--- &0.594 & --- & --- \\
	GLRPE \cite{jung2020global}&\underline{0.502} &  --- & ---& 0.591 & --- &  ---\\
	RSGN \cite{zhao2021reconstructive}&0.450&0.457&0.440&0.601&0.611&\textbf{0.600}\\\hline
	VJMHT (Ours) &   \textbf{0.506} &\textbf{0.517} &  \textbf{0.464} &  \textbf{0.609} & \underline{0.619}   & 0.589  \\ \hline
\end{tabular}
\end{table*}

\noindent \textbf{F-measure.}
F-measure is widely used in previous works to compare video summarization methods. Given a video summary $V_s$ and its corresponding ground truth summary $V_{gt}$, the precision $P$ and recall $R$ are computed as follows,
\begin{equation}
P=\frac{\left| V_s\cap V_{gt}\right| }{\left| V_s\right|},R=\frac{\left| V_s\cap V_{gt}\right| }{\left| V_{gt}\right|},
\end{equation}
\imp{where $|\cdot|$ denotes the number of frames in the sequence.} The F-measure $F$ is the harmonic average of $P$ and $R$, \textit{i.e.},
\begin{equation}
F=\frac{2PR}{P+R}.
\end{equation}
A higher F-measure indicates the summary greatly overlaps with the ground truth while containing less redundancy. 

\noindent \textbf{Rank-based Evaluation.}
It has been pointed out in \cite{otani2019rethinking} that 1) the distribution of shot lengths has a great impact on F-measure; 2) randomly generated summaries sometimes achieve considerable F-measure. In this case, F-measure alone is not sufficient enough to evaluate video summarization. \imp{In this paper, we also adopt the rank-based evaluation \cite{otani2019rethinking} to measure the performance of different methods. Specifically, given the predicted frame-level importance scores and the ground truth ones, two rank correlation coefficients, Kendall’s $\tau$ and Spearman’s $\rho$, are computed.}

\subsection{Comparisons with Existing Methods}

We compare the proposed VJMHT with several existing methods using F-measure on SumMe and TVSum. Among the compared methods, dppLSTM \cite{zhang2016video}, H-RNN \cite{zhao2017hierarchical} and HSA-RNN \cite{zhao2018hsa} use bi-LSTM to model temporal dependencies, while our method exploits the Transformer to capture global dependencies; SUM-GAN \cite{mahasseni2017unsupervised} and ACGAN \cite{he2019unsupervised} perform unsupervised video summarization in GAN, while our method maximizes the similarity between the video and its summary by Transformer-based video representation reconstruction; SUM-FCN \cite{rochan2018video}, GLRPE \cite{jung2020global} and RSGN \cite{zhao2021reconstructive} capture the global dependencies by FCN, attention mechanism and graph network, respectively, while our method models temporal dependencies hierarchically.
Since the Transformer-based video representation reconstruction is introduced in our method, VJMHT can be trained in an unsupervised manner, where $\mathcal{L}_{sup}$ in Eq. (\ref{ul}) is discarded. \textbf{The unsupervised version of VJMHT is denoted as VJMHT$\bm{_{uns}}$}. The results are shown in TABLE \ref{res1}. Note that the methods above VJMHT$_{uns}$ are unsupervised methods or weakly supervised ones, while the methods below VJMHT$_{uns}$ are supervised methods.

\begin{table*}[htbp]
\caption{The comparison results (rank correlation coefficients) on SumMe and TVSum. The methods above VJMHT$_{\bm{uns}}$ are unsupervised methods or weakly supervised ones, while the methods below VJMHT$_{\bm{uns}}$ are supervised methods. The results of randomly generated and human-generated summaries are reported as well. The best and the second-best results are in \textbf{bold} and \underline{underlined}, respectively.}\label{res2}
\center
\begin{tabular}{l|cc|cc}
	\hline
	\multirow{2}{*}{Methods} & \multicolumn{2}{c|}{SumMe} & \multicolumn{2}{c}{TVSum} \\
	& Kendall’s $\tau$    & Spearman’s $\rho$   & Kendall’s $\tau$      & Spearman’s $\rho$    \\ \hline
	Random&     0.000         &      0.000       &   0.000       &     0.000    \\
	Human&     0.205         &     0.213        &      0.177        &    0.204     \\
	\hline
	SUM-GAN \cite{mahasseni2017unsupervised}	&     ---         &  ---           &             0.024& 0.032            \\
	WS-HRL \cite{chen2019weakly}	&    ---          &    ---      &  0.078& \underline{0.116}           \\ 
	DR-DSN \cite{zhou2018deep}&  0.047   &   0.048    &   0.020      &   0.026     \\
RSGN$_{uns}$ \cite{zhao2021reconstructive}&  0.071  &  0.073    &     0.048   &   0.052   \\
	\hline
	VJMHT$_{uns}$ (Ours)&         0.061     &         0.063    &      0.070        &       0.075      \\ 
	\hline
	dppLSTM \cite{zhang2016video}&   ---       &   ---     &    0.042      &    0.055     \\
	CSNet$_{sup}$ \cite{jung2019discriminative}		&       ---       &    ---         &             0.025& 0.034           \\ 
	GLRPE \cite{jung2020global}	&      ---        &      ---       &              0.070& 0.091          \\ 
	\new{SumGraph \cite{park2020sumgraph}}&   \new{---}  &   \new{---}   &    \new{\underline{0.094}}    & \new{\textbf{0.138}}      \\
	HSA-RNN \cite{zhao2018hsa}&  0.064       &     0.066   &     0.082     &   0.088      \\
RSGN \cite{zhao2021reconstructive}&   \underline{0.083}  &     \underline{0.085}  & 0.083     &   0.090     \\
	\hline
	VJMHT (Ours)&      \textbf{0.106}        &        \textbf{0.108}     &      \textbf{0.097}        &       0.105    \\ 
	\hline
\end{tabular}
\end{table*}

As shown in TABLE \ref{res1}, VJMHT$_{uns}$ is comparable to the unsupervised methods and some supervised ones. Specifically, VJMHT$_{uns}$ performs remarkably in the canonical setting and the augmented setting on SumMe. Besides, it achieves the best performance in the augmented settings on TVSum among unsupervised methods. Therefore, by Transformer-based video reconstruction, our model is capable of identifying the important shots without supervision.

In terms of the supervised methods, VJMHT achieves the best results in three settings on SumMe, and the canonical setting on TVSum. Moreover, it outperforms most methods in other settings on TVSum. Although some methods achieve results close to (or better than) ours in certain settings, \textbf{our method performs well CONSISTENTLY in three settings on two datasets}. For example, SUM-FCN achieves good results as ours in the augmented setting on SumMe, but its results in the canonical/transfer setting are inferior; re-SEQ2SEQ is better than VJMHT in the augmented setting on TVSum, while VJMHT greatly outperforms it on SumMe. None of other methods achieve as consistently good performance as ours. In summary, in terms of F-measure, VJMHT performs remarkably on two datasets compared to the state-of-the-art methods.

\begin{table*}[htbp]
\caption{\imp{The results (F-measure) of ablation studies. \textit{CS}, \textit{HT} and \textit{REC} are the abbreviations for video co-summarization, the hierarchical Transformer and Transformer-based  reconstruction, respectively.} The best and the second-best results are in \textbf{bold} and \underline{underlined}, respectively.}
\label{abl1}
\center
\begin{tabular}{ccc|ccc|ccc}
	\hline
	\multirow{2}{*}{\textit{CS}} & \multirow{2}{*}{\textit{HT}} & \multirow{2}{*}{\textit{REC}} & \multicolumn{3}{c|}{SumMe} & \multicolumn{3}{c}{TVSum} \\
	&   &   & Canonical & Augmented & Transfer & Canonical & Augmented & Transfer\\ \hline
	 \multicolumn{3}{c|}{Baseline}  &  0.485 & 0.496  & 0.445  & 0.595  & 0.608 & 0.561 \\\hline
	\checkmark &   &   &  0.483& 0.498 &  0.440 &  0.592& 0.598 & 0.558 \\
	 & \checkmark  &   & 0.494  &0.501   &  0.454 &  0.601 &  0.614 & 0.562  \\
	  &   &  	\checkmark & 0.488 & 0.499 &  0.451 & 0.598 & 0.612 & 0.565 \\\hline
	\checkmark & \checkmark &   & \underline{0.501}  & 0.503  & 0.457  &  0.607 & \underline{0.615}  &  \underline{0.571} \\
	& \checkmark  & \checkmark  & 0.496 & \underline{0.507}  & \underline{0.461}  & \underline{0.608} & 0.614  & 0.566 \\\hline
	\checkmark & \checkmark & \checkmark &  \textbf{0.506} & \textbf{0.517}  &  \textbf{0.464} &  \textbf{0.609 }& \textbf{0.619}  &  \textbf{0.583} \\ \hline
\end{tabular}
\end{table*}

\subsection{Rank-based Evaluation}

Complementary to F-measure, rank-based evaluation is utilized in recent works \cite{jung2020global,jung2019discriminative}. We also adopt it to compare our method with existing ones, as well as the randomly generated and human-generated summaries. The comparison results are shown in TABLE \ref{res2}. Note that the results of the human summary are computed using the leave-one-out approach as in \cite{otani2019rethinking}.

As shown in TABLE \ref{res2}, VJMHT$_{uns}$ outperforms most unsupervised methods and some supervised methods, which means the Transformer-based video representation reconstruction models the relative importance among shots reasonably. With the supervision of human annotations, the performance of VJMHT is improved significantly and surpasses most methods on two datasets in terms of both Kendall’s $\tau$ and Spearman’s $\rho$ correlation coefficients. We conclude that the predicted frame-wise importance scores are well consistent with human annotations.

\subsection{Ablation Study}

In this section, we perform extensive ablation studies on the proposed VJMHT. Their exist three crucial components in VJMHT: video co-summarization by the video joint modelling (\textit{CS}), the hierarchical Transformer (\textit{HT}) and Transformer-based video representation reconstruction (\textit{REC}). To verify the effectiveness of each of them, we construct simplifications of VJMHT as in TABLE \ref{abl1}. The baseline model and the models without the hierarchical structure are constructed by a plain Transformer to encode all the frames simultaneously. The model with only video joint modelling performs frame-level information aggregation across similar videos. The model with only Transformer-based video representation reconstruction uses the encoded special token in the plain Transformer as the video representation. The model with all components is the proposed VJMHT.

As shown in TABLE \ref{abl1}, the baseline model performs considerably, which verifies the effectiveness of Transformer for video summarization. However, with the frame-level video joint modelling, the performance drops slightly. We speculate the reason is that frame-level information aggregation involves too much redundancy and the attention in the model is dispersed. By modeling videos hierarchically, the performance in different settings on two datasets improves significantly, which means the hierarchical structure is more proper for videos than plain sequential models. With the frame-level video representation reconstruction, the performance is improved with respect to the baseline. With the shot-level video joint modelling, the model captures the semantic correlations across videos, so the performance in the canonical setting on the small-scale SumMe is raised obviously. With the hierarchical structure and Transformer-based reconstruction, the results in the transfer setting on two datasets are improved remarkably. We believe the reason is that the reconstruction takes predominant effect when there exists a huge domain gap between the training set and the testing set. In conclusion, the proposed components are of great inportance for VJMHT.

\begin{figure*}[tbp]
\centering
\includegraphics[width=\textwidth]{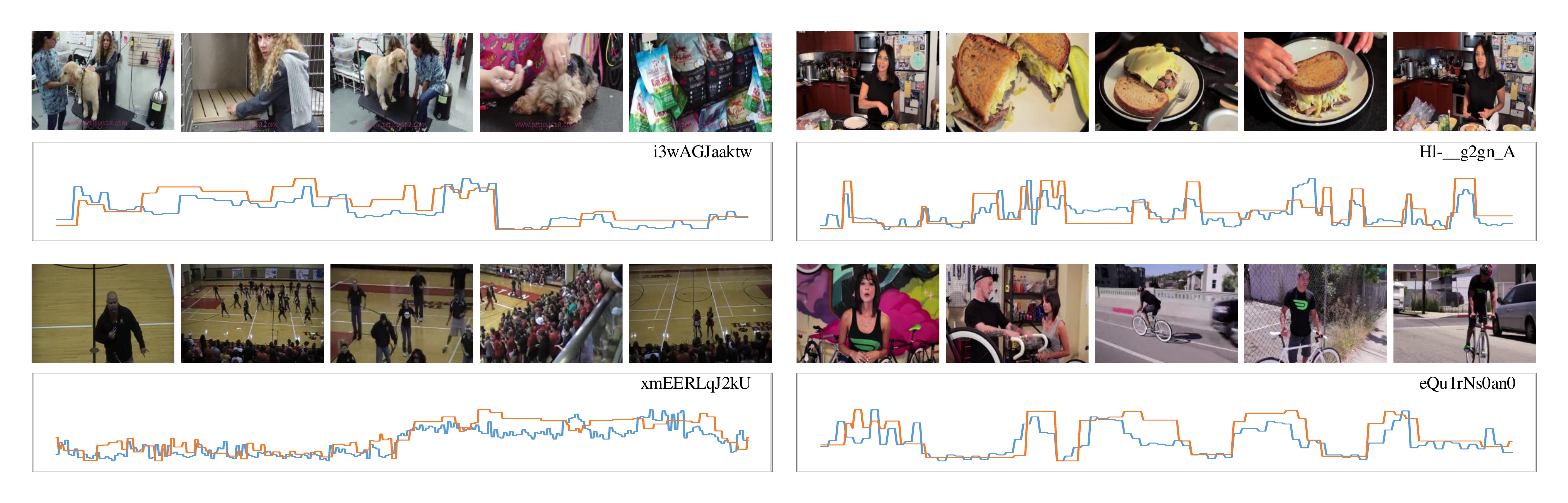}
\caption{The generated summaries and the predicted scores by VJMHT of four videos from TVSum. The keyframes in the first row are sampled from the summaries generated by our method. As for the curves in the second row, the \textcolor[RGB]{91,155,213}{blue} lines depict the ground truth scores, while the \textcolor[RGB]{237,125,49}{orange} lines depict the predicted ones. The video names are on the upper right.}
\label{curve}
\end{figure*}

\subsection{Visualization}

Some generated summaries as well as the predicted scores by VJMHT are illustrated in Fig. \ref{curve}. As we can see from the keyframes (sampled from the summaries generated by our method), the main content of the original videos is contained in the summaries. By only skimming through the summaries, a viewer can infer the main content in the video. In terms of the predicted scores (blue line), they fit the ground truth scores (orange line) well, which means the proposed method can capture the temporal dependencies and measure the importance of shots accurately.

\subsection{Sensitivity Analysis}
\label{fa}

In this subsection, we conduct sensitivity analysis on the proposed method in the canonical setting. Specifically, we study the impact of the number of layers in Transformer, the number of heads in the multi-head attention mechanism, and the dimension of shot embeddings. To eliminate the impact of co-summarization and reconstruction, the video joint modelling and Transformer-based video representation reconstruction are discarded in the those experiments. \new{Besides, we also conduct experiments to demonstrate the impact of the number of videos and the sampling modes of videos for joint modeling, the impact of Transformer-based frame-level modelling, and the quantitative performance of video joint modelling. All settings are the same as mentioned in Section \ref{id} unless specified otherwise.}

\begin{table}[tbp]
\caption{The F-measure of different numbers of layers in the F-Transformer, as well as the inference time on TVSum.}
\label{fa1}
\center
\begin{tabular}{cccc}
\hline
\#Layers & SumMe & TVSum & Runtime (s) \\\hline
1            &   0.481    &   0.589&\textbf{12.49}    \\
2            &   0.494    &   0.601&14.65    \\
3            &   \textbf{0.495}    &   \textbf{0.605}&18.32   \\\hline
\end{tabular}
\end{table}

\begin{table}[tbp]
\caption{The F-measure of different numbers of layers in the S-Transformer, as well as the inference time on TVSum.}
\label{fa2}
\center
\begin{tabular}{cccc}
\hline
\#Layers & SumMe & TVSum & Runtime (s) \\\hline
1            &      0.472&0.593&\textbf{13.67}       \\
2            &      \textbf{0.498}&0.598&14.18      \\
3            &      0.494&\textbf{0.601}&14.65      \\
4            &      0.490&0.597&15.35      \\\hline
\end{tabular}
\end{table}

\subsubsection{Impact of the Number of Layers in Transformer}
The number of layers in Transformer are crucial to the performance and efficiency. Generally, with the increase in the number of layers, the representation ability of the model is improved, but it takes more time to train and test the model. Since our model consists of two Transformers (F-Transformer and S-Transformer), we study the impact of the number of layers in F-Transformer and S-Transformer.

TABLE \ref{fa1} shows the impact of the number of layers in the F-Transformer. As we can see from the results, the one-layer F-Transformer achieves the worst results on two datasets, which means single-layer structure is insufficient to capture the dependencies between frames in each shot. With more layers, the F-measure increases considerably. However, when more layers are equipped, the improvement is less significant, and the efficiency decreases greatly. In this case, we use the two-layer F-Transformer in our model.

TABLE \ref{fa2} shows the impact of the number of layers in the S-Transformer. As we can see from the results, the S-Transformer with two layers achieves the best result on SumMe, while the S-Transformer with three layers achieves the best result on TVSum. Besides, the performance drops on both datasets when more layers are added, which implies over-fitting. Moreover, the number of layers in the S-Transformer has moderate impact on the inference time. We believe the reason is that the numbers of shots in videos are too small to have great influence on the computational complexity of Transformer which is quadratic to the sequence length. Therefore, we set the number of layers in the S-Transformer to 3 in VJMHT.

\begin{table}[tbp]
\caption{The F-measure of different numbers of heads in the F-Transformer.}
\label{fa3}
\center
\begin{tabular}{ccc}
\hline
\#Heads & SumMe & TVSum  \\\hline
1            &   0.491    &   0.599   \\
2            &   \textbf{0.494}    &   0.601  \\
4            &   0.493    &   \textbf{0.604}  \\\hline
\end{tabular}
\end{table}

\subsubsection{Impact of the Number of Heads}

Different to the traditional attention mechanism, the multi-head attention mechanism first transforms the features into several subspaces, and then applies the attention mechanism in different subspaces, respectively. The final output of the multi-head attention mechanism the transformed concatenation of the attention outputs in different subspaces. By this means, Transformer captures the global dependencies from different aspects. We conduct experiments to study the impact of the number of heads in the F-Transformer and S-Transformer, respectively.

TABLE \ref{fa3} shows the impact of the number of heads in the F-Transformer. As we can see from the results, the number of heads in the F-Transformer has little impact on the performance on two datasets. We speculate the reason is that the frames in a shot are semantically similar, so a simple attention mechanism is powerful enough to capture their dependencies. 

TABLE \ref{fa4} shows the impact of the number of heads in the S-Transformer. As we can see from the results, the S-Transformer with the single-head attention mechanism achieves the worst performance on two datasets, which means the correlations between shots are complicated and the simple attention mechanism is not capable of modelling such correlations. With more heads, the performance is improved significantly. Specifically, the S-Transformer with two heads achieves the best result on SumMe, while the S-Transformer with three heads achieves the best result on TVSum. We speculate that the difference between the two datasets in contents accounts for the inconsistency of the impact of number of heads on performance. \imp{Taking into consideration both the F-Transformer and the S-Transformer, the head number in the multi-head mechanism is set to 2.}

\begin{table}[tbp]
\caption{The F-measure of different numbers of heads in the S-Transformer.}
\label{fa4}
\center
\begin{tabular}{ccc}
\hline
\#Heads & SumMe & TVSum  \\\hline
1            &0.476&   0.589    \\
2            &\textbf{0.494}&   0.601    \\
4            &0.490&   \textbf{0.605}  \\\hline
\end{tabular}
\end{table}

\subsubsection{Impact of the Dimension of Shot Embeddings}
\begin{table}[tbp]
\caption{The F-measure of different dimensions of shot embeddings.}
\label{fa5}
\center
\begin{tabular}{ccc}
\hline
Dimension & SumMe & TVSum  \\\hline
256            &  0.459     &   0.585    \\
512            &   \textbf{0.494}    &   \textbf{0.601}    \\
1,024            &  0.472     &  0.596    \\\hline
\end{tabular}
\end{table}

TABLE \ref{fa5} shows the impact of the dimension of shot embeddings. As we can see from the results, the model with the small dimension of shot embeddings achieves poor F-measure, which indicates the representation ability of the model is limited. When increasing the dimension to 512, the performance reaches the best on two datasets. However, with the large dimension of 1,024, the performance drops. We assume over-fitting occurs in this setting. In this case, we choose to set the dimension of shot embeddings to 512 throughout the experiments.

\subsubsection{\new{Impact of the Number of Videos for Joint Modelling}}

\new{To demonstrate the impact of the number of videos for joint modelling, we conduct experiments on TVSum, and the validation results are shown in TABLE \ref{fa6}. Note that the results of the model without video joint modelling are shown in the first row of the table, where the number of videos is 1. As shown in the table, the rank correlation coefficients first increase and then decrease, where the peak is reached at two videos. We also find that too more videos for joint modelling may compromise the performance compared to the model without video joint modelling. One possible reason is that with the increase in the number of videos, the number of shots for joint modelling increases significantly and the attention for each shot is greatly diminished. Considering both performance and memory use, we choose to use two videos for joint modelling in this work.}

\begin{table}[tbp]
\caption{\new{The results of different numbers of videos for joint modelling on TVSum.}}
\label{fa6}
\center
\begin{tabular}{ccc}
\hline
\#Videos  & Kendall's $\tau$ & Spearman's $\rho$ \\ \hline
1  & 0.088 & 0.095 \\
2  & \textbf{0.097} & \textbf{0.105} \\
3  & 0.092 & 0.099 \\
4  & 0.085 & 0.091 \\ \hline
\end{tabular}
\end{table}


\subsubsection{\new{Impact of the Sampling Modes of Videos}}

\new{To demonstrate the impact of the semantic relation between the videos for joint modelling, we conduct experiments on TVSum, where 1) the videos are randomly selected from the training set instead of the same cluster (Random), and 2) the videos are randomly selected from different clusters (Inter-Cluster). The results are shown in TABLE \ref{fa7}, including the model without video joint modelling (w/o VJM) as baseline. As shown in TABLE \ref{fa7}, the results of random selection and inter-cluster selection are similar to those of the training without video joint modelling, which indicates that video joint modelling using unrelated videos hardly compromises the baseline performance. Additionally, by using semantically similar videos for video joint modelling during training (Intra-Cluster), the performance is improved significantly compared with the baseline, which means the related videos are beneficial to learning common importance patterns in videos.}

\begin{table}[tbp]
\caption{\new{The results of different the sampling mode on TVSum.}}
\label{fa7}
\center
\begin{tabular}{ccc}
\hline
Mode & Kendall's $\tau$ & Spearman's $\rho$ \\ \hline
w/o VJM & 0.088 & 0.095 \\
Random & 0.085 & 0.903 \\
Inter-Cluster & 0.084 & 0.903 \\
Intra-Cluster & \textbf{0.097} & \textbf{0.105} \\ \hline
\end{tabular}
\end{table}

\begin{table}[tbp]
\caption{\new{The results of different frame-level modelling techniques on TVSum.}}
\label{fa8}
\center
\begin{tabular}{cccc}
\hline
Techniques & F-measure & Kendall's $\tau$ & Spearman's $\rho$ \\ \hline
C3D & 0.572 & 0.055 & 0.060 \\
Transformer & \textbf{0.609} & \textbf{0.097} & \textbf{0.105}\\\hline
\end{tabular}
\end{table}

\subsubsection{\new{Impact of Transformer-based Frame-level Modelling}}

\new{In our method, the frame-level dependencies within each shot are captured by Transformer. To demonstrate whether 3D-CNN \cite{tran2015learning} is effective to capture the frame-level dependencies, we conduct experiments by replacing the F-Transformer with 3D-CNN to obtain the shot representations. Specifically, for each shot, we utilize the feature extractor pre-trained on Sports-1M dataset \cite{karpathy2014large} to extract C3D (conv5) layer features. Since C3D only accepts clips of 16 frames as input, we first segment the shot into several 16-frame clips, and then extract features from each clip. The final semantic representation of each shot is computed as the average feature of the clips within it. Besides, spatial average pooling is also applied to decrease the feature dimension. Finally, a representation of 512D is extracted from each shot, which is used for cross-video inter-shot dependency modelling as proposed in our paper. The results of using C3D for frame-level dependencies capturing on TVSum are reported in TABLE \ref{fa8}. As we can see from the results, once we use the C3D features as the representations of shots, the performance of video summarization drops significantly compared with our method. The results indicate that the proposed F-Transformer is not substitutable in terms of frame-level correlations capturing. The inferior performance of C3D is expected, because C3D captures only the local dependencies within a clip, which is insufficient to model the complex correlations among all frames in the shot. In conclusion, the proposed F-Transformer is crucial for frame-level modelling and cannot be replaced with the C3D network.}

\subsubsection{\new{Qualitative Evaluation of Video Joint Modelling}}

\begin{figure}[tbp]
\centering
\includegraphics[width=\columnwidth]{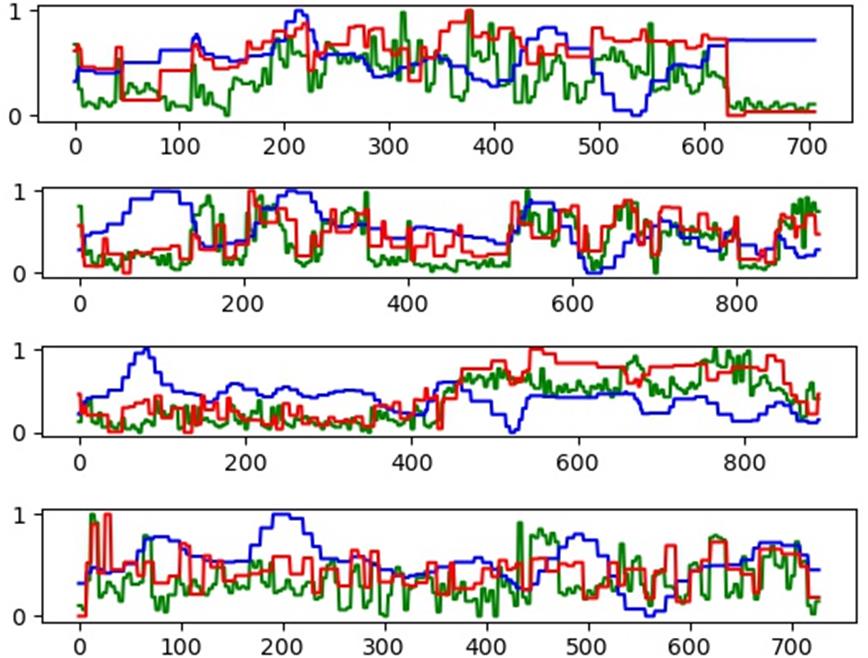}
\caption{\new{The visualization of the predicted important scores by the model without video joint modelling (\textcolor[RGB]{0,0,255}{blue} line) and the model with video joint modelling (\textcolor[RGB]{255,0,0}{red} line). The \textcolor[RGB]{0,255,0}{green} lines show the ground truth scores.}}
\label{vjmab}
\end{figure}

\new{Apart from evaluating the proposed video joint modelling quantitatively, we also visualize the predicted important scores to qualitatively demonstrate the effectiveness of video joint modelling. The visualization results of four videos from the TVSum dataset are shown in Fig. \ref{vjmab}. As we can see from the figure, the model with video joint modelling (red line) fits the ground truth (green line) better than the model without video joint modelling (blue line) does. The results indicate that video joint modelling is beneficial for measuring the relative importance among different shots in videos. In conclusion, video joint modelling is effective in both quantitative and qualitative evaluations.}

\section{Conclusion}
\label{con}

In this paper, we propose Video Joint Modelling based on Hierarchical Transformer (VJMHT) for co-summarization. Specifically, VJMHT consists of two layers of Transformer to model the videos hierarchically. The first Transformer encodes the frames in each shot, while the second Transformer captures the global dependencies among all shots. To exploit the semantic similarity across videos, video joint modelling is designed and incorporated in the second Transformer, where the shots of similar videos are encoded simultaneously. Moreover, Transformer-based video representation reconstruction is introduced in the training process to minimize the semantic distance between the summary and the video. \imp{The experiment results have proved the effectiveness of the proposed components, and the superiority of VJMHT in terms F-measure and rank correlation coefficients.}

\section*{Acknowledgements}
This research was undertaken using the LIEF HPC-GPGPU Facility. This Facility was established with the assistance of LIEF Grant LE170100200. This work was also supported by Grant 2022ECR008. MG was supported by DE210101624.

\ifCLASSOPTIONcaptionsoff
  \newpage
\fi



\bibliographystyle{IEEEtran}
\bibliography{ref}

\begin{thebibliography}{10}
\providecommand{\url}[1]{#1}
\csname url@samestyle\endcsname
\providecommand{\newblock}{\relax}
\providecommand{\bibinfo}[2]{#2}
\providecommand{\BIBentrySTDinterwordspacing}{\spaceskip=0pt\relax}
\providecommand{\BIBentryALTinterwordstretchfactor}{4}
\providecommand{\BIBentryALTinterwordspacing}{\spaceskip=\fontdimen2\font plus
\BIBentryALTinterwordstretchfactor\fontdimen3\font minus
  \fontdimen4\font\relax}
\providecommand{\BIBforeignlanguage}[2]{{%
\expandafter\ifx\csname l@#1\endcsname\relax
\typeout{** WARNING: IEEEtran.bst: No hyphenation pattern has been}%
\typeout{** loaded for the language `#1'. Using the pattern for}%
\typeout{** the default language instead.}%
\else
\language=\csname l@#1\endcsname
\fi
#2}}
\providecommand{\BIBdecl}{\relax}
\BIBdecl

\bibitem{mitra2016bayesian}
A.~Mitra, S.~Biswas, and C.~Bhattacharyya, ``Bayesian modeling of temporal
  coherence in videos for entity discovery and summarization,'' \emph{IEEE
  transactions on pattern analysis and machine intelligence}, vol.~39, no.~3,
  pp. 430--443, 2016.

\bibitem{sun2016summarizing}
M.~Sun, A.~Farhadi, B.~Taskar, and S.~Seitz, ``Summarizing unconstrained videos
  using salient montages,'' \emph{IEEE transactions on pattern analysis and
  machine intelligence}, vol.~39, no.~11, pp. 2256--2269, 2016.

\bibitem{zhang2016video}
K.~Zhang, W.-L. Chao, F.~Sha, and K.~Grauman, ``Video summarization with long
  short-term memory,'' in \emph{European conference on computer vision}.\hskip
  1em plus 0.5em minus 0.4em\relax Springer, 2016, pp. 766--782.

\bibitem{mahasseni2017unsupervised}
B.~Mahasseni, M.~Lam, and S.~Todorovic, ``Unsupervised video summarization with
  adversarial lstm networks,'' in \emph{Proceedings of the IEEE conference on
  Computer Vision and Pattern Recognition}, 2017, pp. 202--211.

\bibitem{zhou2018deep}
K.~Zhou, Y.~Qiao, and T.~Xiang, ``Deep reinforcement learning for unsupervised
  video summarization with diversity-representativeness reward,'' in
  \emph{Proceedings of the AAAI Conference on Artificial Intelligence},
  vol.~32, no.~1, 2018.

\bibitem{zhao2017hierarchical}
B.~Zhao, X.~Li, and X.~Lu, ``Hierarchical recurrent neural network for video
  summarization,'' in \emph{Proceedings of the 25th ACM international
  conference on Multimedia}, 2017, pp. 863--871.

\bibitem{yuan2019spatiotemporal}
Y.~Yuan, H.~Li, and Q.~Wang, ``Spatiotemporal modeling for video summarization
  using convolutional recurrent neural network,'' \emph{IEEE Access}, vol.~7,
  pp. 64\,676--64\,685, 2019.

\bibitem{zhao2018hsa}
B.~Zhao, X.~Li, and X.~Lu, ``Hsa-rnn: Hierarchical structure-adaptive rnn for
  video summarization,'' in \emph{Proceedings of the IEEE conference on
  computer vision and pattern recognition}, 2018, pp. 7405--7414.

\bibitem{fajtl2018summarizing}
J.~Fajtl, H.~S. Sokeh, V.~Argyriou, D.~Monekosso, and P.~Remagnino,
  ``Summarizing videos with attention,'' in \emph{Asian Conference on Computer
  Vision}.\hskip 1em plus 0.5em minus 0.4em\relax Springer, 2018, pp. 39--54.

\bibitem{wei2019sequence}
Z.~Wei, B.~Wang, M.~Hoai, J.~Zhang, X.~Shen, Z.~Lin, R.~Mech, and D.~Samaras,
  ``Sequence-to-segments networks for detecting segments in videos,''
  \emph{IEEE transactions on pattern analysis and machine intelligence}, 2019.

\bibitem{hussain2021comprehensive}
T.~Hussain, K.~Muhammad, W.~Ding, J.~Lloret, S.~W. Baik, and V.~H.~C.
  de~Albuquerque, ``A comprehensive survey of multi-view video summarization,''
  \emph{Pattern Recognition}, vol. 109, p. 107567, 2021.

\bibitem{li2017general}
X.~Li, B.~Zhao, and X.~Lu, ``A general framework for edited video and raw video
  summarization,'' \emph{IEEE Transactions on Image Processing}, vol.~26,
  no.~8, pp. 3652--3664, 2017.

\bibitem{zhao2021reconstructive}
B.~Zhao, H.~Li, X.~Lu, and X.~Li, ``Reconstructive sequence-graph network for
  video summarization,'' \emph{IEEE Transactions on Pattern Analysis and
  Machine Intelligence}, 2021.

\bibitem{chu2015video}
W.-S. Chu, Y.~Song, and A.~Jaimes, ``Video co-summarization: Video
  summarization by visual co-occurrence,'' in \emph{Proceedings of the IEEE
  Conference on Computer Vision and Pattern Recognition}, 2015, pp. 3584--3592.

\bibitem{vaswani2017attention}
A.~Vaswani, N.~Shazeer, N.~Parmar, J.~Uszkoreit, L.~Jones, A.~N. Gomez,
  {\L}.~Kaiser, and I.~Polosukhin, ``Attention is all you need,''
  \emph{Advances in neural information processing systems}, vol.~30, pp.
  5998--6008, 2017.

\bibitem{de2011vsumm}
S.~E.~F. De~Avila, A.~P.~B. Lopes, A.~da~Luz~Jr, and
  A.~de~Albuquerque~Ara{\'u}jo, ``Vsumm: A mechanism designed to produce static
  video summaries and a novel evaluation method,'' \emph{Pattern Recognition
  Letters}, vol.~32, no.~1, pp. 56--68, 2011.

\bibitem{ngo2003automatic}
C.-W. Ngo, Y.-F. Ma, and H.-J. Zhang, ``Automatic video summarization by graph
  modeling,'' in \emph{Proceedings Ninth IEEE International Conference on
  Computer Vision}.\hskip 1em plus 0.5em minus 0.4em\relax IEEE, 2003, pp.
  104--109.

\bibitem{mei20142}
S.~Mei, G.~Guan, Z.~Wang, M.~He, X.-S. Hua, and D.~D. Feng, ``L 2, 0
  constrained sparse dictionary selection for video summarization,'' in
  \emph{2014 IEEE international conference on multimedia and expo
  (ICME)}.\hskip 1em plus 0.5em minus 0.4em\relax IEEE, 2014, pp. 1--6.

\bibitem{rochan2018video}
M.~Rochan, L.~Ye, and Y.~Wang, ``Video summarization using fully convolutional
  sequence networks,'' in \emph{Proceedings of the European Conference on
  Computer Vision (ECCV)}, 2018, pp. 347--363.

\bibitem{ji2019video}
Z.~Ji, K.~Xiong, Y.~Pang, and X.~Li, ``Video summarization with attention-based
  encoder--decoder networks,'' \emph{IEEE Transactions on Circuits and Systems
  for Video Technology}, vol.~30, no.~6, pp. 1709--1717, 2019.

\bibitem{huang2019novel}
C.~Huang and H.~Wang, ``A novel key-frames selection framework for
  comprehensive video summarization,'' \emph{IEEE Transactions on Circuits and
  Systems for Video Technology}, vol.~30, no.~2, pp. 577--589, 2019.

\bibitem{park2020sumgraph}
J.~Park, J.~Lee, I.-J. Kim, and K.~Sohn, ``Sumgraph: Video summarization via
  recursive graph modeling,'' in \emph{Computer Vision--ECCV 2020: 16th
  European Conference, Glasgow, UK, August 23--28, 2020, Proceedings, Part XXV
  16}.\hskip 1em plus 0.5em minus 0.4em\relax Springer, 2020, pp. 647--663.

\bibitem{zhang2021i2net}
W.~Zhang, B.~Wang, S.~Ma, Y.~Zhang, and Y.~Zhao, ``I2net: Mining intra-video
  and inter-video attention for temporal action localization,''
  \emph{Neurocomputing}, vol. 444, pp. 16--29, 2021.

\bibitem{wang2021exploring}
B.~Wang, X.~Zhang, and Y.~Zhao, ``Exploring sub-action granularity for weakly
  supervised temporal action localization,'' \emph{IEEE Transactions on
  Circuits and Systems for Video Technology}, 2021.

\bibitem{han2020mining}
M.~Han, Y.~Wang, X.~Chang, and Y.~Qiao, ``Mining inter-video proposal relations
  for video object detection,'' in \emph{European Conference on Computer
  Vision}.\hskip 1em plus 0.5em minus 0.4em\relax Springer, 2020, pp. 431--446.

\bibitem{wang2020contrastive}
N.~Wang, W.~Zhou, and H.~Li, ``Contrastive transformation for self-supervised
  correspondence learning,'' \emph{arXiv preprint arXiv:2012.05057}, 2020.

\bibitem{zhang2019scan}
R.~Zhang, J.~Li, H.~Sun, Y.~Ge, P.~Luo, X.~Wang, and L.~Lin, ``Scan:
  Self-and-collaborative attention network for video person
  re-identification,'' \emph{IEEE Transactions on Image Processing}, vol.~28,
  no.~10, pp. 4870--4882, 2019.

\bibitem{carion2020end}
N.~Carion, F.~Massa, G.~Synnaeve, N.~Usunier, A.~Kirillov, and S.~Zagoruyko,
  ``End-to-end object detection with transformers,'' in \emph{European
  Conference on Computer Vision}.\hskip 1em plus 0.5em minus 0.4em\relax
  Springer, 2020, pp. 213--229.

\bibitem{zhu2020deformable}
X.~Zhu, W.~Su, L.~Lu, B.~Li, X.~Wang, and J.~Dai, ``Deformable detr: Deformable
  transformers for end-to-end object detection,'' \emph{arXiv preprint
  arXiv:2010.04159}, 2020.

\bibitem{girdhar2019video}
R.~Girdhar, J.~Carreira, C.~Doersch, and A.~Zisserman, ``Video action
  transformer network,'' in \emph{Proceedings of the IEEE/CVF Conference on
  Computer Vision and Pattern Recognition}, 2019, pp. 244--253.

\bibitem{gavrilyuk2020actor}
K.~Gavrilyuk, R.~Sanford, M.~Javan, and C.~G. Snoek, ``Actor-transformers for
  group activity recognition,'' in \emph{Proceedings of the IEEE/CVF Conference
  on Computer Vision and Pattern Recognition}, 2020, pp. 839--848.

\bibitem{zhang2018retrospective}
K.~Zhang, K.~Grauman, and F.~Sha, ``Retrospective encoders for video
  summarization,'' in \emph{Proceedings of the European Conference on Computer
  Vision (ECCV)}, 2018, pp. 383--399.

\bibitem{he2016deep}
K.~He, X.~Zhang, S.~Ren, and J.~Sun, ``Deep residual learning for image
  recognition,'' in \emph{Proceedings of the IEEE conference on computer vision
  and pattern recognition}, 2016, pp. 770--778.

\bibitem{ba2016layer}
J.~L. Ba, J.~R. Kiros, and G.~E. Hinton, ``Layer normalization,'' \emph{arXiv
  preprint arXiv:1607.06450}, 2016.

\bibitem{potapov2014category}
D.~Potapov, M.~Douze, Z.~Harchaoui, and C.~Schmid, ``Category-specific video
  summarization,'' in \emph{European conference on computer vision}.\hskip 1em
  plus 0.5em minus 0.4em\relax Springer, 2014, pp. 540--555.

\bibitem{devlin2018bert}
J.~Devlin, M.-W. Chang, K.~Lee, and K.~Toutanova, ``Bert: Pre-training of deep
  bidirectional transformers for language understanding,'' \emph{arXiv preprint
  arXiv:1810.04805}, 2018.

\bibitem{tay2020efficient}
Y.~Tay, M.~Dehghani, D.~Bahri, and D.~Metzler, ``Efficient transformers: A
  survey,'' \emph{arXiv preprint arXiv:2009.06732}, 2020.

\bibitem{gygli2014creating}
M.~Gygli, H.~Grabner, H.~Riemenschneider, and L.~Van~Gool, ``Creating summaries
  from user videos,'' in \emph{European conference on computer vision}.\hskip
  1em plus 0.5em minus 0.4em\relax Springer, 2014, pp. 505--520.

\bibitem{song2015tvsum}
Y.~Song, J.~Vallmitjana, A.~Stent, and A.~Jaimes, ``Tvsum: Summarizing web
  videos using titles,'' in \emph{Proceedings of the IEEE conference on
  computer vision and pattern recognition}, 2015, pp. 5179--5187.

\bibitem{smeaton2006evaluation}
A.~F. Smeaton, P.~Over, and W.~Kraaij, ``Evaluation campaigns and trecvid,'' in
  \emph{Proceedings of the 8th ACM international workshop on Multimedia
  information retrieval}, 2006, pp. 321--330.

\bibitem{jung2019discriminative}
Y.~Jung, D.~Cho, D.~Kim, S.~Woo, and I.~S. Kweon, ``Discriminative feature
  learning for unsupervised video summarization,'' in \emph{Proceedings of the
  AAAI Conference on Artificial Intelligence}, vol.~33, no.~01, 2019, pp.
  8537--8544.

\bibitem{szegedy2015going}
C.~Szegedy, W.~Liu, Y.~Jia, P.~Sermanet, S.~Reed, D.~Anguelov, D.~Erhan,
  V.~Vanhoucke, and A.~Rabinovich, ``Going deeper with convolutions,'' in
  \emph{Proceedings of the IEEE conference on computer vision and pattern
  recognition}, 2015, pp. 1--9.

\bibitem{russakovsky2015imagenet}
O.~Russakovsky, J.~Deng, H.~Su, J.~Krause, S.~Satheesh, S.~Ma, Z.~Huang,
  A.~Karpathy, A.~Khosla, M.~Bernstein \emph{et~al.}, ``Imagenet large scale
  visual recognition challenge,'' \emph{International journal of computer
  vision}, vol. 115, no.~3, pp. 211--252, 2015.

\bibitem{kingma2014adam}
D.~P. Kingma and J.~Ba, ``Adam: A method for stochastic optimization,''
  \emph{arXiv preprint arXiv:1412.6980}, 2014.

\bibitem{he2019unsupervised}
X.~He, Y.~Hua, T.~Song, Z.~Zhang, Z.~Xue, R.~Ma, N.~Robertson, and H.~Guan,
  ``Unsupervised video summarization with attentive conditional generative
  adversarial networks,'' in \emph{Proceedings of the 27th ACM International
  Conference on Multimedia}, 2019, pp. 2296--2304.

\bibitem{zhao2019property}
B.~Zhao, X.~Li, and X.~Lu, ``Property-constrained dual learning for video
  summarization,'' \emph{IEEE transactions on neural networks and learning
  systems}, vol.~31, no.~10, pp. 3989--4000, 2019.

\bibitem{rochan2019video}
M.~Rochan and Y.~Wang, ``Video summarization by learning from unpaired data,''
  in \emph{Proceedings of the IEEE/CVF Conference on Computer Vision and
  Pattern Recognition}, 2019, pp. 7902--7911.

\bibitem{chen2019weakly}
Y.~Chen, L.~Tao, X.~Wang, and T.~Yamasaki, ``Weakly supervised video
  summarization by hierarchical reinforcement learning,'' in \emph{Proceedings
  of the ACM Multimedia Asia}, 2019, pp. 1--6.

\bibitem{jung2020global}
Y.~Jung, D.~Cho, S.~Woo, and I.~S. Kweon, ``Global-and-local relative position
  embedding for unsupervised video summarization,'' in \emph{European
  Conference on Computer Vision, ECCV 2020}.\hskip 1em plus 0.5em minus
  0.4em\relax Springer, 2020.

\bibitem{otani2019rethinking}
M.~Otani, Y.~Nakashima, E.~Rahtu, and J.~Heikkila, ``Rethinking the evaluation
  of video summaries,'' in \emph{Proceedings of the IEEE/CVF Conference on
  Computer Vision and Pattern Recognition}, 2019, pp. 7596--7604.

\bibitem{tran2015learning}
D.~Tran, L.~Bourdev, R.~Fergus, L.~Torresani, and M.~Paluri, ``Learning
  spatiotemporal features with 3d convolutional networks,'' in
  \emph{Proceedings of the IEEE international conference on computer vision},
  2015, pp. 4489--4497.

\bibitem{karpathy2014large}
A.~Karpathy, G.~Toderici, S.~Shetty, T.~Leung, R.~Sukthankar, and L.~Fei-Fei,
  ``Large-scale video classification with convolutional neural networks,'' in
  \emph{Proceedings of the IEEE conference on Computer Vision and Pattern
  Recognition}, 2014, pp. 1725--1732.

\end{thebibliography}
%

%

\begin{IEEEbiography}[{\includegraphics[width=1in,height=1.25in,clip,keepaspectratio]{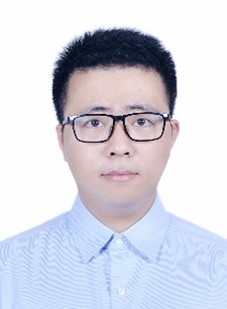}}]{Haopeng Li}
is a Ph.D student in the School of Computing and Information Systems, University of Melbourne. He received his B.S. degree in the School of Science, Northwestern Polytechnical University, and the Master degree in the School of Artificial Intelligence, Optics and Electronics (iOPEN), Northwestern Polytechnical University. His research interests include computer vision, video understanding, and artificial intelligence.
\end{IEEEbiography}

\begin{IEEEbiography}[{\includegraphics[width=1in,height=1.25in,clip,keepaspectratio]{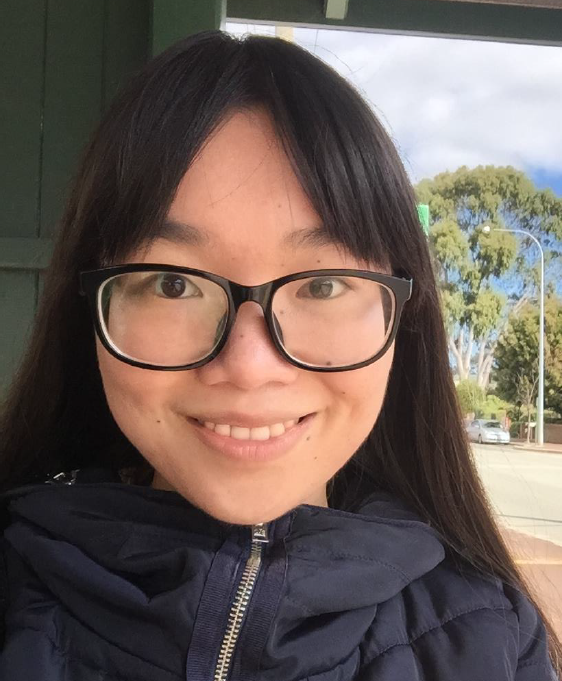}}]{Qiuhong Ke}
received her PhD degree from The University of Western Australia in 2018. She is a Lecturer (Assistant Professor) at Monash University. Before that, she was a Postdoctoral Researcher at Max Planck Institute for Informatics and a Lecturer at University of Melbourne. Her thesis “Deep Learning for Action Recognition and Prediction” has been awarded “Dean’s List-Honourable mention” by The University of Western Australia in 2018. She was awarded “1962 Medal” for her work in video recognition technology by the Australian Computer Society in 2019. She was also awarded Early Career Researcher Award by Australia Pattern Recognition Society in 2020. Her research interests include computer vision and machine learning.
\end{IEEEbiography}

.


\begin{IEEEbiography}[{\includegraphics[width=1in,height=1.25in,clip,keepaspectratio]{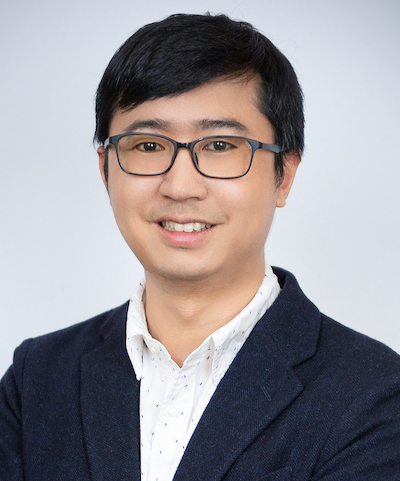}}]{Mingming Gong}
is a Lecturer (Assistant Professor) in data science with the School of Mathematics and Statistics, the University of Melbourne. His research interests include causal reasoning, machine learning, and computer vision. He has authored and co-authored 40+ research papers on top venues such as ICML, NeurIPS, UAI, AISTATS, IJCAI, AAAI, CVPR, ICCV, ECCV, with 10+ oral/spotlight presentations and a best paper finalist at CVPR19. He has served as area chairs of NeurIPS’21 and ICLR’21, senior program committee members of AAAI'19-20, IJCAI’20-21, program committee members of ICML, NeurIPS, UAI, CVPR, ICCV, and reviewers of TPAMI, AIJ, MLJ, TIP, TNNLS, etc. He received the Discovery Early Career Researcher Award from Australian Research Council in 2021.
\end{IEEEbiography}

\begin{IEEEbiography}[{\includegraphics[width=1in,height=1.25in,clip,keepaspectratio]{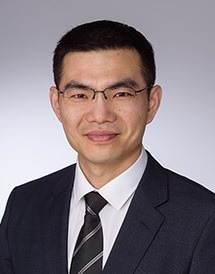}}]{Rui Zhang}
(\url{www.ruizhang.info}) is a visiting professor at Tsinghua University and was a Professor at the University of Melbourne. His research interests include AI and big data, particularly in the areas of recommender systems, knowledge bases, chatbot, and spatial and temporal data analytics. Professor Zhang has won several awards including Future Fellowship by the Australian Research Council in 2012, Chris Wallace Award for Outstanding Research by the Computing Research and Education Association of Australasia in 2015, and Google Faculty Research Award in 2017.
\end{IEEEbiography}




\end{document}